\pdfoutput=1

\documentclass[11pt]{article}

\usepackage[final]{acl}

\usepackage{times}
\usepackage{latexsym}
\usepackage{subcaption}
\renewcommand{\footnotesize}{\scriptsize}
\usepackage{booktabs}
\usepackage{multirow}
\usepackage{graphicx}
\usepackage{float}
\usepackage{amsmath}
\usepackage{amssymb}

\usepackage[T1]{fontenc}

\usepackage[utf8]{inputenc}

\usepackage{microtype}

\usepackage{inconsolata}

\usepackage{graphicx}
\graphicspath{{./figure/}}
\DeclareGraphicsExtensions{.pdf,.jpeg,.png,.jpg}

\title{Infinite Retrieval: Attention Enhanced LLMs in Long-Context Processing}

\author{ Xiaoju Ye\\
  \texttt{yxj2017@gmail.com} \\\And
  Zhichun Wang\\
  \texttt{zcwang@bnu.edu.cn}  \\\And
  Jingyuan Wang\\
  \texttt{202321081040@mail.bnu.edu.cn} \\
}

\begin{document}

\maketitle
\begin{abstract}
Limited by the context window size of Large Language Models(LLMs), handling various tasks with input tokens exceeding the upper limit has been challenging, whether it is a simple direct retrieval task or a complex multi-hop reasoning task. Although various methods have been proposed  to enhance the long-context processing capabilities of LLMs, they either incur substantial post-training costs, or require additional tool modules(e.g.,RAG), or have not shown significant improvement in realistic tasks. Our work observes the correlation between the attention distribution and generated answers across each layer, and establishes the attention allocation aligns with retrieval-augmented capabilities through experiments. Drawing on the above insights, we propose a novel method \textbf{InfiniRetri} that leverages the LLMs's own attention information to enable accurate retrieval across inputs of infinitely length. Our evaluations indicate that InfiniRetri achieves 100\% accuracy in the Needle-In-a-Haystack(NIH) test over \textbf{1M tokens} using a 0.5B parameter model, surpassing other method or larger models and setting  a new \textbf{state-of-the-art(SOTA)}. Moreover, our method achieves significant performance improvements on real-world benchmarks, with a maximum \textbf{288\% improvement}. In addition, InfiniRetri can be applied to any Transformer-based LLMs without additional training and substantially reduces inference latency and compute overhead in long texts. In summary, our comprehensive studies show InfiniRetri's potential for practical applications and creates a paradigm for retrievaling information using LLMs own capabilities under infinite-length tokens. Code will be released in \href{https://github.com/MrYxJ/InfiniRetri}{link}.

\begin{figure}[ht]
  \centering
  \subfloat{\includegraphics[width=\columnwidth]{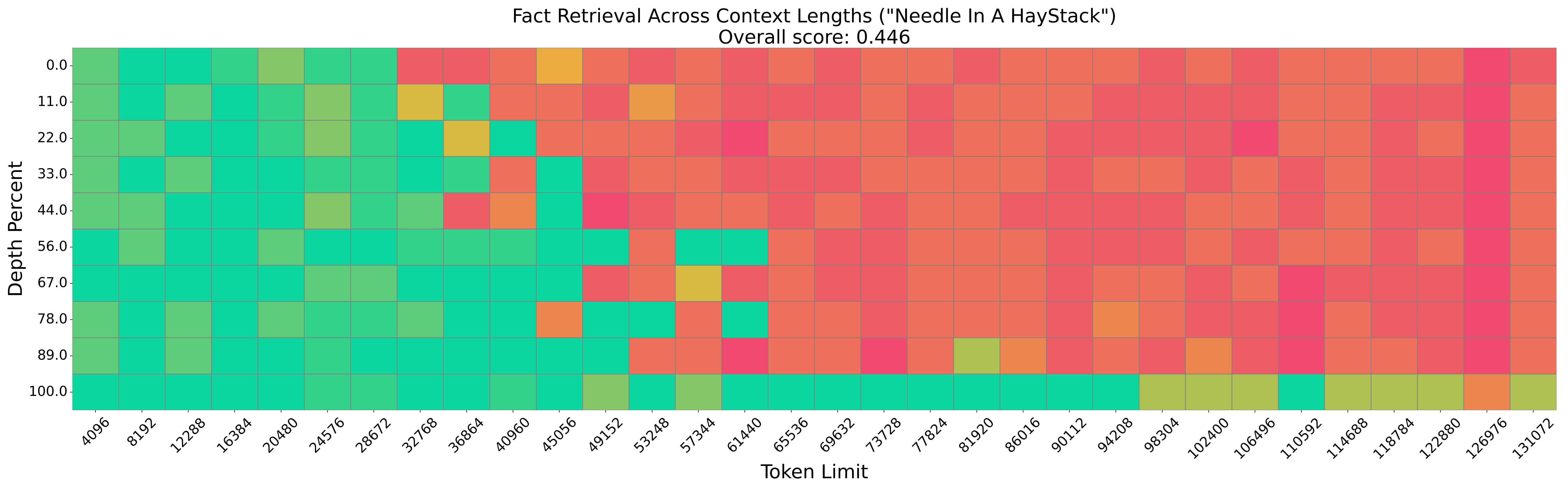}}
  
  \subfloat{\includegraphics[width=\columnwidth]{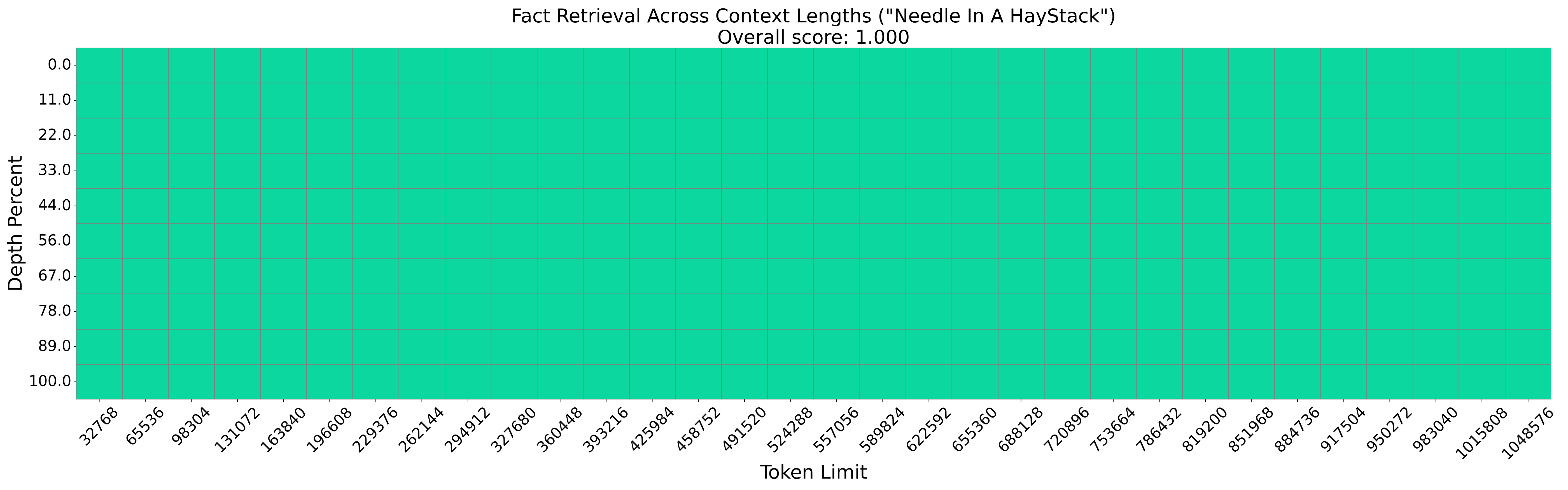}}
 
  \caption{ \label{fig:Qwen0.5B-compare}
     Performance of Qwen2.5-0.5B-Instruct in NIH test with original(Top) and \textbf{InfiniRetri}(Bottom), the maximum length of the accurately haystack from original 32K over to \textbf{1M} tokens.
  }
\end{figure}

\end{abstract}

\section{Introduction}
Large Language Models(LLMs)\citep{achiam2023gpt, touvron2023llama, jiang2023mistral, yang2024qwen2, glm2024chatglm} have been widely integrated into various tasks and applications in the field of Natural Language Process(NLP) and the broader domain of Artificial Intelligence (AI), including NLP's system dialogue\citep{chiang2023vicuna}, document summarization\citep{fabbri2019multi} and code completion\citep{roziere2023code}. Simultaneously, the size of the context window has always been a critical indicator of LLMs's capability in processing the aforementioned tasks\citep{wang2024beyond}, as a larger context window allows LLMs to handle longer input contexts within a single window. Concurrently, the OpenAI's o1\citep{openai-o1} model has utilized Chain-of-Thought(CoT)\citep{wei2022chain} to increase length of reasoning process and thereby improved reasoning capabilities\citep{zhong2024evaluation} for LLMs, which triggers a hot research trend on improving reasoning performance. Meanwhile, combined with the research trend of increasing length of mixed-modality data input by multimodal models\citep{zhang2024mm}, it is anticipated that the length of input into LLMs will continue to increase. Consequently, the ability to handle long-context inputs will continue to be essential and increasingly demanded for LLMs.

As analyzed above, recently many leading LLMs pay more attention to scale up the context window, such as GPT-4\citep{achiam2023gpt} and Llama3\citep{dubey2024llama} and DeepSeekV3\citep{liu2024deepseek} up to context length 128K, Claude-3 up to 200K, Gemini-Pro-1.5\citep{team2024gemini} and Qwen2.5-1M\citep{qwen2.5-1m} up to 1M. Despite impressive progress, the actual results, as described in Ruler\citep{hsieh2024ruler}, do not attain the claimed lengths, and there are  still significant challenges in enhancing ability to handle long-context for LLMs. Let's consider the first question: \textbf{Is it only possible to scale up the context window longer and longer?} Firstly, obtaining a longer context window significantly increases computational costs leads to a significant delay due to quadratic computation of attention\citep{vaswani2017attention}. Secondly, considering the long-tail effect of input length, longer texts appear with lower probability, the current approach of simply scaling up the context window is destined to yield lower and lower benefits of capability. Thirdly, the leading open-sources LLMs are basically apply YaRN\citep{peng2023yarn} for continued and phased training to extend context window\citep{gao2024train}, which incurs prohibitively high costs are inaccessible to the vast majority of researchers.

In contrast, enhancing the capability to handle longer contexts from the perspective of low-cost or even training-free is both economical and more challenging. Previously, methods from this perspective mainly focused on Positional Extrapolation and Interpolation\citep{chen2023extending}, which achieved extrapolation beyond the training length by adjusting the Positional embedding(PE) associated with input tokens. However, beyond PE, further research reveals that the attention mechanism itself causes a sharp performance drop when processing sequences longer than the training context length. Consequently, some methods adopt approaches to divide the input contexts into segments and apply a \textbf{Sliding Window}\citep{jiang2023mistral} to iteratively maintain each segment's lengths within the context window. Based on this \textit{segment context and slide window} in long texts method, it is indeed possible to handle longer even infinite-length tokens without training, methods like StreamingLLM\citep{xiao2023efficient}, which innovative addresses the “attention sink” phenomenon to generate responses of unlimited length. However, the model's effective attention and memory scope retains \textbf{confined within a single context window}, and information spanning beyond multiple windows cannot be effectively processed and aggregated.

To address this issue, recently approach mainly focus on Key-Value(KV) Cache Compression, which attempts to \textbf{break the flow of information} between the different context windows and achieve global information by compressing the previous key and value states embedding in the cache to realize manage memory, such as H2O\citep{zhang2023h2o}, SnapKV\citep{li2024snapkv}, InfLLM\citep{xiao2024infllm}, CAMELoT\citep{he2024camelot}, InfiniPot\citep{kim2024infinipot}, CAKE\citep{qin2025cake}, PyramidKV\citep{cai2024pyramidkv}, DynamicKV\citep{zhou2024dynamickv}\textit{et}. Regrettably, current methods either exhibit limited effectiveness or require extensive training to adapt the new mechanisms. Traditional attention-based transformer LLMs are not pretrained with the capability to compress KV caches, unless they incorporate KV-cache-like mechanisms from the beginning of pretraining, such as the innovations in infini-attention\citep{munkhdalai2024leave} and DeepSeek\cite{liu2024deepseek}, which definitedly incured significant costs.Let's further consider the second question: \textbf{Is there a low-cost method to break the information barriers between different context windows} in handling long context? 

In fact, the industry has already provided an answer in the form of Retrieval-Augmented Generation(RAG)\citep{gao2023retrieval, zhao2024retrieval}, which is a framework composed of two primary components:a retrieval module and a generation module. The retrieval module relies on an external embedding model to retrieve relevant passages from long contexts based on input query, which incurs a common challenge in RAG systems: establishing associations between retrieved information is difficult. Conversely, the attention mechanism in LLMs excels at efficiently establishing associations between different pieces of information during inference. This leads to the third question: \textbf{Why not use retrieval capabilities of LLMs themselves to handle long contexts?}

Based on the above consideration, we innovatively propose a training-free method named InfiniRetri to enhance the long-context capabilities of LLMs. Specifically, our method is inspired by the \textit{segment context and slide window} method and employs an iterative mechanism to achieve the capability to handle unlimited contexts. We then critically analyze the limitations of current mainstream KV cache compression methods and identify that their ineffectiveness is primarily due to the inability to enable LLMs to compress and store past keys and values in a low-cost manner. Fundamentally, we argue that only by breaking down the information barriers between different context windows can LLMs truly enhance their ability to handle long texts. In addition, by observing the attention allocation of LLMs during the inference when answering questions, we innovatively propose that the \textit{attention allocation pattern aligns with retrieval-augmented} capabilities. Based on this insight, our method introduces a novel approach that leverages the LLMs' own attention information rather than relying on external embedding models to improve their long-context capabilities.

Benefiting from the fact that our method can be applied to transformer-based LLMs without training, we conduct comprehensive and comparative experiments on multiple models, including Llama3-8B-instruct, Mistral-7B-Instruct-v0.2, Qwen2-7B-Instruct, \textit{et}. In the Fact Retrieval Across Context Lengths("Needle In A HayStack")\citep{liu2024lost, fu2024data}, InfiniRetri extends the model with only 0.5B parmaters from original 32K length over to 1M tokens(as shown in Figure~\ref{fig:Qwen0.5B-compare}). More remarkably, our method enables that using InfiniRetri method on NIH task can achieve accurate retrieval over an \textbf{infinite length} range, which not only outperforms current mainstream methods but also effectively solves the NIH task. In addition, out method achieves 9 realistic datasets from LongBench\citep{bai2023longbench} surpassed the existing mainstream methods based on KV Cache and even Full KV of models, especially in the Multi-Document QA tasks such as HotpotQA, 2WikiMQA and Musique, where the Qwen2-7B-Instruct using our method achieve a significant improvement of \textbf{369.6\%} on average.

In summary, our main contribution are as follow:
\begin{itemize}
    \item We innovatively propose the concept: \textbf{attention allocation alignment with retrieval-augmented} and successfully leverage it to enhance the long-text processing capabilities of LLMs.
    \item Our method supports \textbf{training-free} application to any Transformer-based LLMs, endowing it with the capability to handle infinitely long contexts.
    \item Unlike RAG, which relies on external embedding model, our method introduces the novel insight of: \textbf{retrieval in attention}, which leverages the inherent capabilities of LLMs to enhance their ability to handle long texts, which may offer new possibilities for the development of RAG and related techniques.
    \item Our method significantly reduces inference latency and computational overhead, excels at handling retrieval and question-answering tasks over massive datasets, which demonstrates substantial practical value in scenarios involving extremely long contexts.
    \item Rather than simply extending context window, we also demonstrate that enhancing the long-text capabilities of LLMs can be achieved through multiple approaches. Future improvements in long-text handling can be achieved by strengthening the model's \textbf{internal capabilities within a smaller context window}, thereby achieving better long-context performance.
\end{itemize}

\section{Related Works}
\subsection{Towards Long Context Window}
With increasing demand for long-text processing, currently leading LLMs are trending towards longer context window lengths in order to enhance their capabilities for handling long texts. For example, DeepseekV3 and Command-R\citep{command} can handle up to 128K, while ChatGLM4 and InternLM2\citep{cai2024internlm2} can process up to 1M tokens. These examples reflect the importance of extending context windows. The above foundation LLMs are primarily continued and phased training by adjusting the base frequency of the RoPE\citep{su2024roformer}, which requires efficient design and substantial computational resources to support additional training on long texts\citep{xiong2023effective}.

Considering the substantial training resources, LongLoRA\citep{chen2023longlora} fine-tuned LLMs to handle long texts through the efficient LoRA\cite{hu2021lora} method. SelfExtend\citep{jin2024llm} constructed bi-level attention information by implementing the position encoding calculation without finetuning to extend context window. However, such training-free methods have limited improvement effects. TransformerFAM\citep{hwang2024transformerfam} innovatively proposed a feedback attention mechanism as working memory to manage the past context in cache. Its operational principle is similar to that of KV Cache Compression methods.

\subsection{KV Cache Compression}
Considering both costs and improvement effects, researchers found that efficient manage of the KV Cache is crucial method to improve the ability of handling long texts, such as H2O\citep{zhang2023h2o}, StreamingLLM\citep{xiao2023efficient}, SnapKV\citep{li2024snapkv}, InfiniPot\citep{kim2024infinipot}, DuoAttention\citep{xiao2024duoattention}, PyramidKV\citep{cai2024pyramidkv}, DynamicKV\citep{zhou2024dynamickv}, CAKE\citep{qin2025cake} \textit{et}.Overall, the fundamental framework of these methods involves designing distinct strategies to retrain a subset of tokens in the cache, thereby reducing computational costs during inference.

Although these methods, including H2O, SnapKV, InfiniPot, PyramidKV and DynamicKV have been incrementally optimized and improved on prior work, their reliance on caching past key-value states fundamentally constrains their performance. No matter how these methods are designed, they cannot match the performance of full key-value (Full KV) caching at a low cost. Compared to Full KV, these methods can only achieve reduce memory and computational costs. In contrast, our method not only reduces costs but also outperforms Full KV in terms of effectiveness. The essence of this result is that the aforementioned methods have not sufficiently leveraged the inherent capabilities of LLMs to design strategies for caching tokens.

Specifically, methods such as H2O, InfiniPot, PyramidKV, and SnapKV all considered the allocation of attention scores as a feature in their method design. Starting from SnapKV, through PyramidKV to DynamicKV, there has been in-depth research proposing that the distribution of LLM’s attention scores follows a specific pattern. SnapKV also emphasizes a viewpoint similar to ours, suggesting that "LLMs know what you are looking for before generation". However, since these methods all cache tokens at the granularity of individual token, they have not fully leveraged this phenomenon to achieve the desired improvements in long-context processing.

\section{Observations}
\label{sec:observation}

\begin{figure*}[ht]
  \centering
  \subfloat[Layer 0\label{fig:layer0}]{\includegraphics[width=2.1\columnwidth]{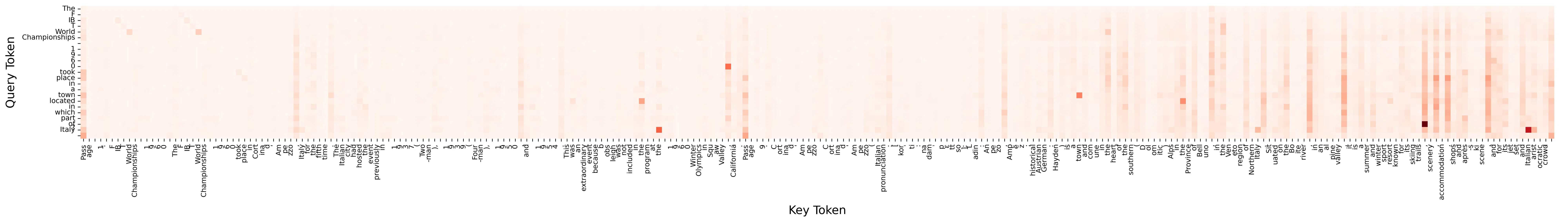}}
  
  \subfloat[Layer 1\label{fig:layer1} ]{\includegraphics[width=2.1\columnwidth]{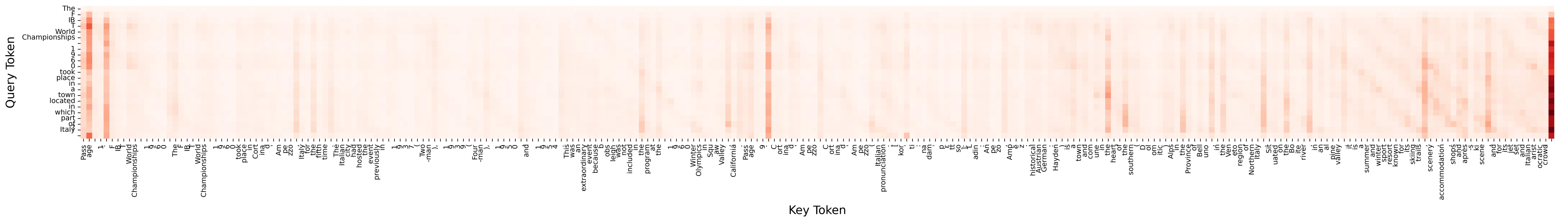}}

  \subfloat[Layer 27(last)\label{fig:layer27}]{\includegraphics[width=2.1\columnwidth]{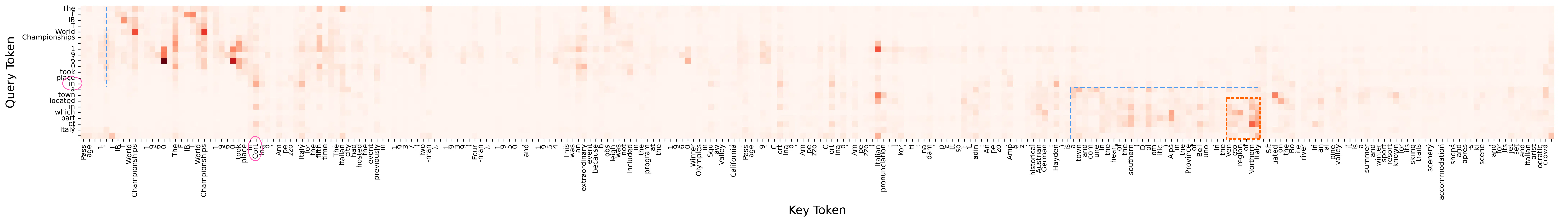}}

  \caption{ \label{fig:obervation}
    Visual 0, 1, 2, 26, 27 layers of Attention Scores Heatmap from using Qwen2-7B-Instruct inference in a QA Sample
  }
\end{figure*}

In this section, we present how we derive the insight:"\textit{attention allocation pattern aligns with retrieval-augmented}" by observing the distribution of attention scores during the LLMs inference, and why this insight plays a crucial role in enhancing the long-context processing capabilities of our proposed method.

In the paper by SnapKV\citep{li2024snapkv}, it was demonstrated that the attention allocation of LLMs during inference exhibits a stable pattern in the Query-Key matrix. However, this observation only confirms that LLMs focus on specific regions within the context based on the query and generate corresponding responses. It does not guarantee that the regions attended to by the LLMs' attention mechanism contain the correct answers. In other words, we need to further verify whether the attention allocation pattern of LLMs can accurately locate the correct answers in the context based on the query. This verification is the key determinant of whether the pattern is effective.

To this end, we specifically selected a Question-Answer(QA) task dataset for testing, focusing on the distribution of the attention scores, that is, the Query-Key matrix where the Question text serves as the Query Token and the context(Answer in here) text serves as the Key Token. For example, as shown in Figure~\ref{fig:obervation}, we extracted a sample segment from HotpotQA, we use the \textbf{Visual Attention Allocation} function \footnote{This function was developed in InfiniRetri.} of InfiniRetri to illustrate the distribution of the attention scores for this QA sample across the 0, 1 and 27(last) layers of the LLMs during the inference. Our observation of Figure~\ref{fig:obervation} reveals that while the attention score distributions in the shallow layers of LLMs appear irregularity, those \textbf{in layers closer to the output exhibit increasingly distinct patterns}. This QA sample requires two-step reasoning. The Question is \textit{"The FIBT World Championships 1960 took place in a town located in which part of Italy ?"}. To answer correctly, one must integrate information from two distinct parts of the context. Text1 is \textit{"FIBT World Championships 1960 took place in Cortina d'Ampezzo"}, and Text2 is \textit{"in the Veneto region of Northern Italy"}, where Text2 is correct answer. As illustrated in Figures~\ref{fig:layer0} and~\ref{fig:layer1}, in the initial layers (layers 0, 1), the Query Token's attention to the correct answer in Key Token is not pronounced. However, as depicted in the last layer(as Figure~\ref{fig:layer27}), the attention score distribution clearly indicates that the LLM successfully focuses on the regions most two relevant to the Query within the corresponding Key, which are exactly Text1 and Text2. Specifically, as shown in Figure~\ref{fig:layer27}, the LLM can accurately attend to distinct regions in the context corresponding to different emphases in the Question. Specifically, for the phrases: \textit{"FIBT World Championships 1960 took place"} and \textit{"located in which part of Italy?"} in the Question, the LLM correctly focuses on the respective regions \textit{"...... in Cortina d'Ampezzo"} and \textit{"...... of Northern Italy"}. Notably, as highlighted by the red circle in Figure~\ref{fig:layer27}, the LLM precisely attends to "\textit{Cort}" in Key Token and "\textit{in}" token in the Query Token, which demonstrating it \textbf{token-level} accuracy of attention mechanism. 

\begin{figure}[htbp]
  \centering
  \subfloat{\includegraphics[width=1.5\columnwidth, height=0.8\linewidth, keepaspectratio]{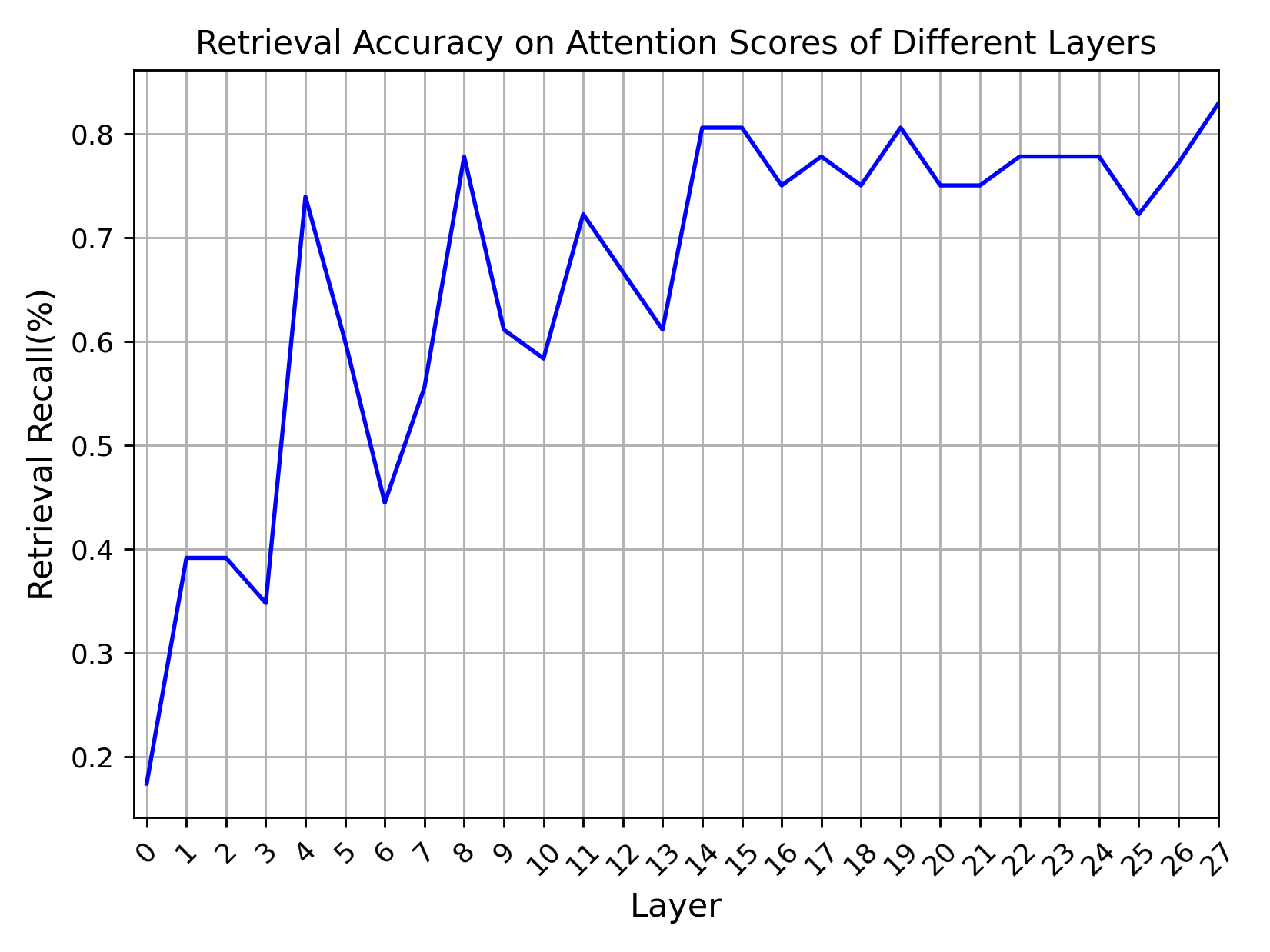}} 
  \caption{The retrieval accuracy on LLMs each layers \label{fig:all_layers_line_chart}}
\end{figure}

To further understand the performance differences across layers of this pattern, we tested the accuracy of retrieving answers based on the attention distribution from all layers. As illustrated in Figure~\ref{fig:all_layers_line_chart}, the closer the layer is to the output layer, the more pronounced the effect of enhancing this pattern. Further, the retrieval accuracy reaches a local peak at layers 14 and 15. Simultaneously, the visualization\footnote{We visual all layers attention scores of this QA sample in Appendix~\ref{appendix:Example of visual all layers}} of attention score distributions across all layers for QA Sample reveals that the correct answer regions in layers 14 and 15 are assigned significantly higher attention scores by the model, which demonstrates a striking correlation between the two phenomena. Above all, this confirms that mining the attention allocation of LLMs enhances their ability to retrieve answers through questions, indicating the \textbf{attention pattern aligned with retrieval-augmented}. This insight also provides guiding suggestions for designing our method. 

\begin{figure*}[htbp]
  \centering
  \subfloat{\includegraphics[width=2.25\columnwidth, keepaspectratio]{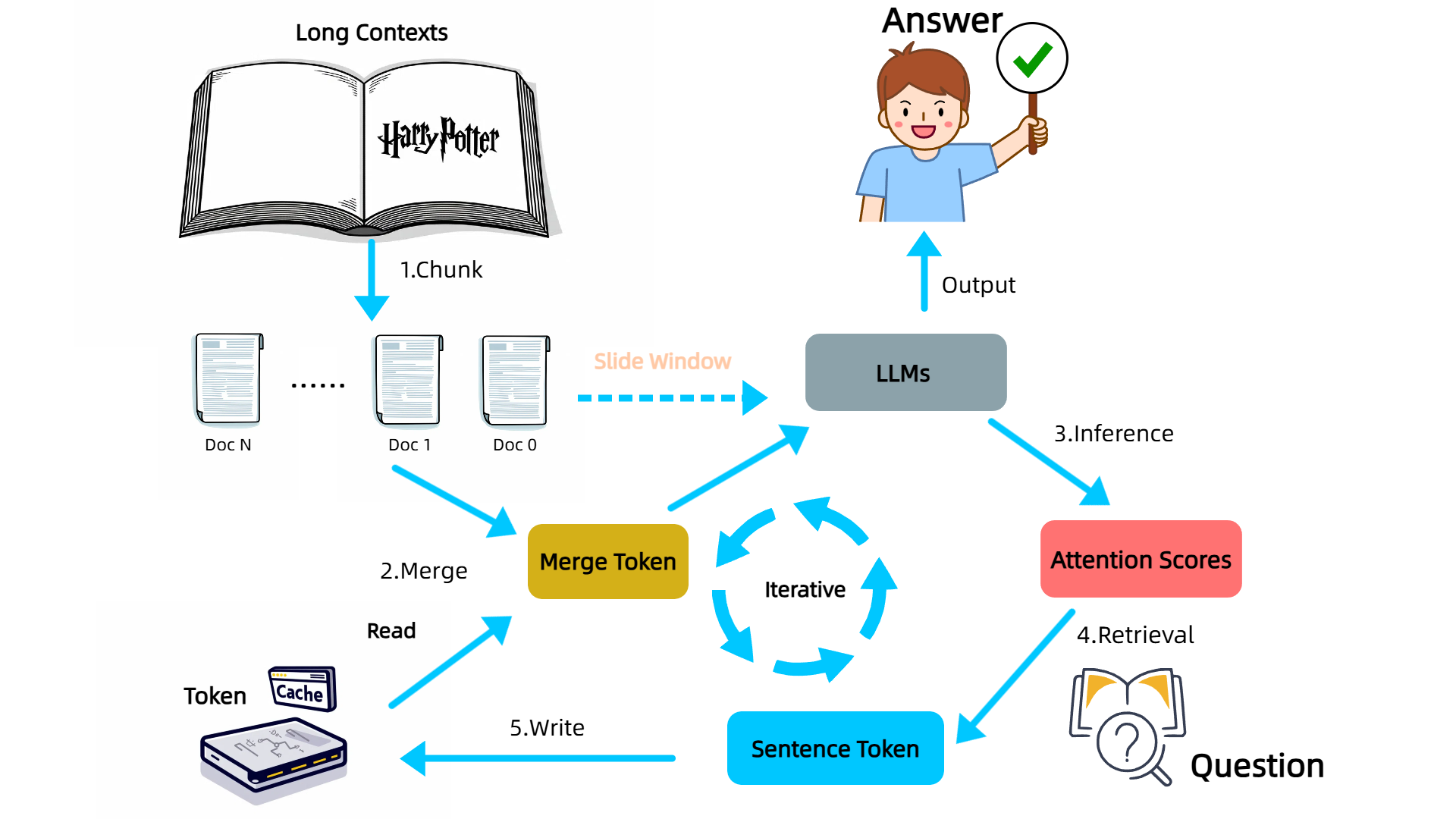}} 
  \caption{Entire Workflow of Our Method InfiniRetri for Enhancing Long-Context Processing in LLMs
\label{fig:InfiniRetri_Workflow}}
\end{figure*}

\section{Method}
Then, how can we apply this pattern to process the long texts that exceed the context window and genuinely enhance long-text capabilities? In this section, we introduce InfiniRetri by dividing it into three subsections, which collectively demonstrate the application of this pattern to improve long-context capabilities in LLMs. As illustrated in Figure~\ref{fig:InfiniRetri_Workflow}, which depicts the entire workflow of our method, the five main steps are introduced in Sections~\ref{sec:4.1} to ~\ref{sec:4.3}. Specifically, Step 1(chunk), Step 2(merge), and Step 3(inference) are detailed in Section~\ref{sec:4.1}; Step 4(retrieval), which is the most critical part of our method, is described in Section~\ref{sec:4.2}; and Step 5(cache) is covered in Section~\ref{sec:4.3}.

\subsection{Segment and Slide Window}
\label{sec:4.1}

\begin{figure}[ht]
  \centering
  \subfloat{\includegraphics[width=0.9\columnwidth, height=0.6\linewidth]{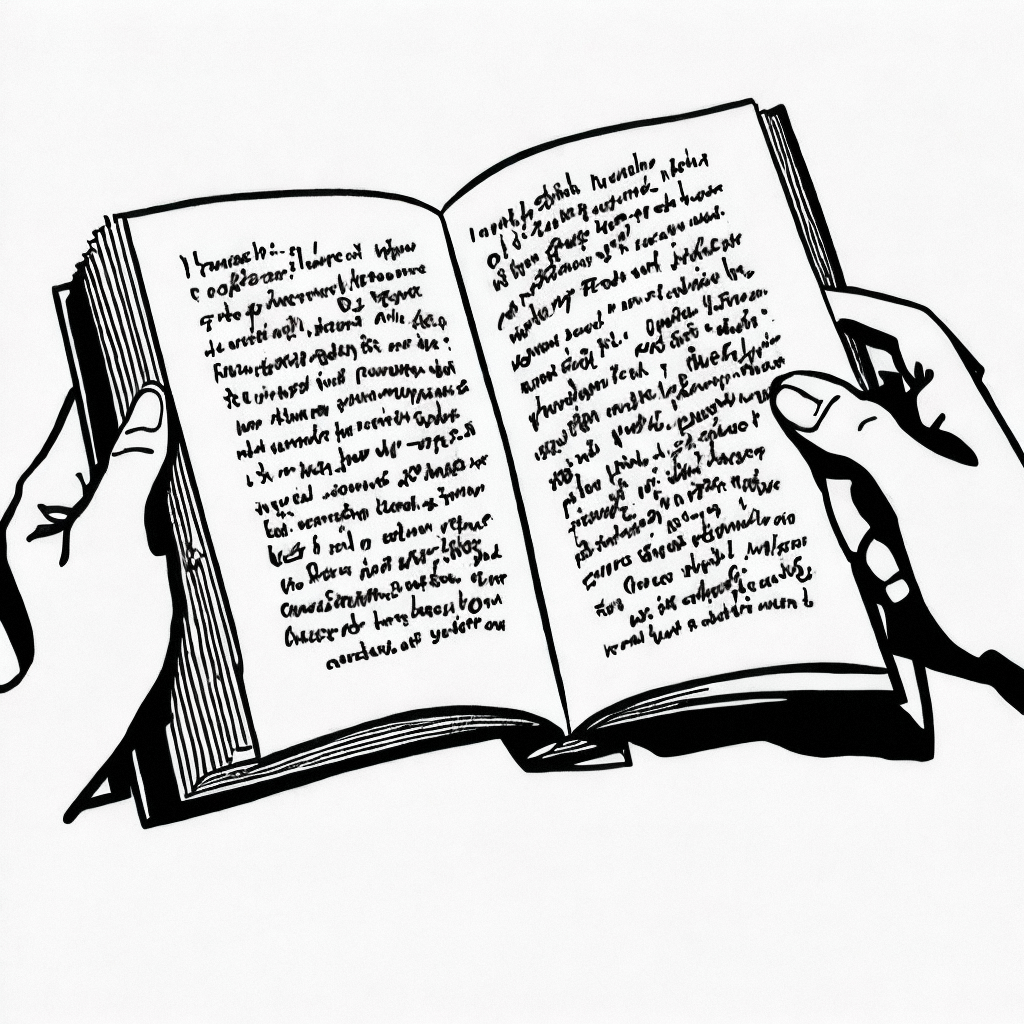}}
  \caption{Humans are limited by their field of vision, but can read the entire book page by page.\label{fig:compare_method}}
\end{figure}

Our method, inspired by the human process of reading books, addresses the challenge of processing texts that exceed the context window of LLMs. Despite the limited field of vision that allows us to see only one page at a time, we can still read and comprehend an entire book by reading each page sequentially. In this process, the brain acts like a \textbf{cache} to retain and integrate information from each page through memory, thereby acquiring the full content of the book. Similarly, our method segments the entire text into continuous chunks. This chunking process is akin to that in RAG, but instead of processing each chunk in parallel,  we iteratively process each segment doc in order. This approach of preserving the order information\citep{yu2024defense} aligns more closely with human reading habits.

Specifically, as shown in Figure~\ref{fig:InfiniRetri_Workflow}, the Step 1(Chunk) segments the entire long context into approximately equal-length documents based on sentence boundaries,  determined by method's parameter \textit{Chunk Size}\footnote{It has great impact of our method, for more details see Appendix~\ref{Appendix:method_parameters}}. These documents are then sequentially merged with the tokens previously retained in the Cache to form complete input sequences, referred to as MergeToken, which are fed into the LLMs. Our method follows a similar iterative approach to Slide Window Attention(SWA), processing each text segment in a sequential manner. However, our handling of the cache is fundamentally different. Instead of using the traditional cache that stores past key-value states at each layer, we repurpose the caching concept by storing past token IDs. As depicted in Figure~\ref{fig:InfiniRetri_Workflow} Step 2(Merge), our method merge these cached token IDs with the current segment’s tokens before feeding them into the model. This merging process replaces the need for merging past key-value states during model inference. Consequently, the Step 3(Inference), which involves LLM inference, employs the standard attention mechanism instead of SWA, for $h$-th head attention scores formulated as follows:

\begin{equation}
\label{eq:standard_euqation}
A^h = \text{softmax}\left( \frac{Q^h \cdot (K^h)^\top}{\sqrt{d^h}} \right),
\end{equation}
where $A \in \mathbb{R}^{n \times m}$ denote the matrix representing the queries and keys, where $n$ is the number of queries and $m$ is the number of keys.

\subsection{Retrieval In Attention}
\label{sec:4.2}
As concluded in Section~\ref{sec:observation}, attention allocation patterns facilitate LLMs in accurately locating the correct answers in context tokens within a single context window based on the question tokens. if we consistently apply this pattern across each inference within a sliding window framework, theoretically, it enable the LLMs to reason over the entire context with a constant query, aligning with the fundamental process of human reading.This is similar to the acknowledged learning strategy of \textbf{"reading with questions in mind"}, where we use the question as an anchor to 
 sequentially consolidate pertinent information within a length that the LLMs can handle. Thus, the LLMs' ability to precisely retrieve the most contextually relevant text based on the question is fundamental to the effectiveness of our method. The crux lies in devising a token retrieval strategy and algorithm predicated on the distribution of attention scores.

 Drawing from experiments in Section~\ref{sec:observation}, we selected the last layer of the Multi-Head attention and then aggregate the attention by summing all heads (as shown in Eq.~\ref{eq:sum multi_head}) to explore a method capable of accurately determining the model's areas of most focus. Benefiting from visualization of all attentions scores, we keenly observed that information relevant to the answer is typically composed of consecutive tokens, i.e., at the phrase words granularity. This finding is consistent with our experiments in Figure~\ref{fig:layer27}, which confirmed that LLMs achieve attention precision at the token-level. Consequently, the operation we aim to design involves computing the sum of attention scores for each token and its adjacent tokens in the 2D matrix of attention scores. This computed result will serve as a new feature for ranking in the subsequent retrieval process. Upon preceding in-depth analysis, we identified that this operation is equivalent to a \textbf{1D convolution} using a kernel $K$ filled with ones. Therefore, for the $i$-th token in query and the $j$-th token in keys, the featrue importance $t_{ij}$ is formulated as shown in Eq.~\ref{eq:token_importance}:
 
\begin{equation}
\label{eq:sum multi_head}
A = \sum_{h=1}^{H} A^{h}
\end{equation}

\begin{equation}
\label{eq:token_importance}
t_{ij} = (A * K)_{i, j} = \sum_{u=0}^{k-1} A_{i, j+u}
\end{equation}
where $k$ denotes the size of the 1D-convolutional kernel, which is also the value of parameter \textit{Phrase Token Num} in our method.\footnote{This parameter determines how many adjacent tokens are aggregated when computing the importance feature in attention scores. Details in Appendix~\ref{Appendix:method_parameters}} We then perform a summation along the columns of the matrix $t_{ij}$. The resulting score for each token in the context represents its overall importance, computed as the cumulative sum of its score across all question tokens.Finally, the importance scores $s_i$ for the $i$-th token in the context is computed as shown in Eq.~\ref{eq:final_token_importance}.

\begin{equation}
\label{eq:final_token_importance}
s_{i} =  \sum_{j=0}^{n-1} t_{j, i}
\end{equation}

We select the \textit{Top-K}\footnote{This parameter determines how many token selected from context token to retrieval. Details are provided in Appendix~\ref{Appendix:method_parameters}} context's tokens with the highest importance scores and write their all sentence tokens in the cache. This process is formulated as follows:
\begin{equation}
\begin{aligned}
\text{Top-} K(\mathbf{v}) &= \{ \text{arg}\max_{i} v_i \mid i \in \{1, 2, \ldots, n\} \\
& \quad \ \ \ and \  v_i \geq \text{nth-largest}(v, k) \}
\end{aligned}
\end{equation}

\subsection{Cache Sentence Token}
\label{sec:4.3}
Our method's use of the cache during inference is fundamentally different from the original approach. Instead of directly utilizing the cache, we adopt the concept of using it to store past context information. The specific differences are twofold:
\begin{itemize}
    \item Our method caches token IDs outside the model instead of the past key-value states from each layer. Specifically, we do not use the past key-value cache during inference. Instead, we merge past context information with the current input before each inference.
    \item Our method employs phrase-level features for retrieval, and the cache stores sentence-level tokens that contain the Top-K tokens. Specifically, we store entire sentences rather than individual tokens in the cache.
\end{itemize}

In fact, these two innovative modifications are precisely why our method outperforms prior KV cache methods in enhancing the long-text capabilities of LLMs without finetuning. Our method does not aim to compress tokens in the cache but rather retains relevant contextual information at the sentence level, this is because we consider sentences to be the minimal complete semantic units that ensure the LLMs' understanding instead of single tokens. Lastly,  during the iterative inference of each segment by the LLMs, the intermediate results retained in the cache are dynamically determined by the combination of previously retained tokens and the current segment's input. As a result, these intermediate results are subject to relative changes throughout the process.

\section{Experiments}
In this section, we conduct comprehensive comparisons to validate the effectiveness of our method, including section~\ref{sec:experiment_NIH} demonstrates that our method achieves SOTA results on the NIH task, section~\ref{sec:experiment_LongBench} presents a detailed comparison of InfiniRetri and baseline methods on LongBenchV1,V2. Meanwhile, for readability, we present implementation details and additional experiments in Appendix~\ref{experiment:implementation},~\ref{experiment:ablation_study},~\ref{experiment:reduce}, including the dataset and parameters used in experiments, ablation studies comparing the differences between our method and prior KV cache methods, the effectiveness in reducing latency and overhead.

\subsection{Visualization on NIH Task}
\label{sec:experiment_NIH}

\begin{figure*}[!htbp]
  \centering
  \subfloat[FullKV\label{fig:nih_llama_full}]{\includegraphics[width=2\columnwidth, height=0.2\linewidth]{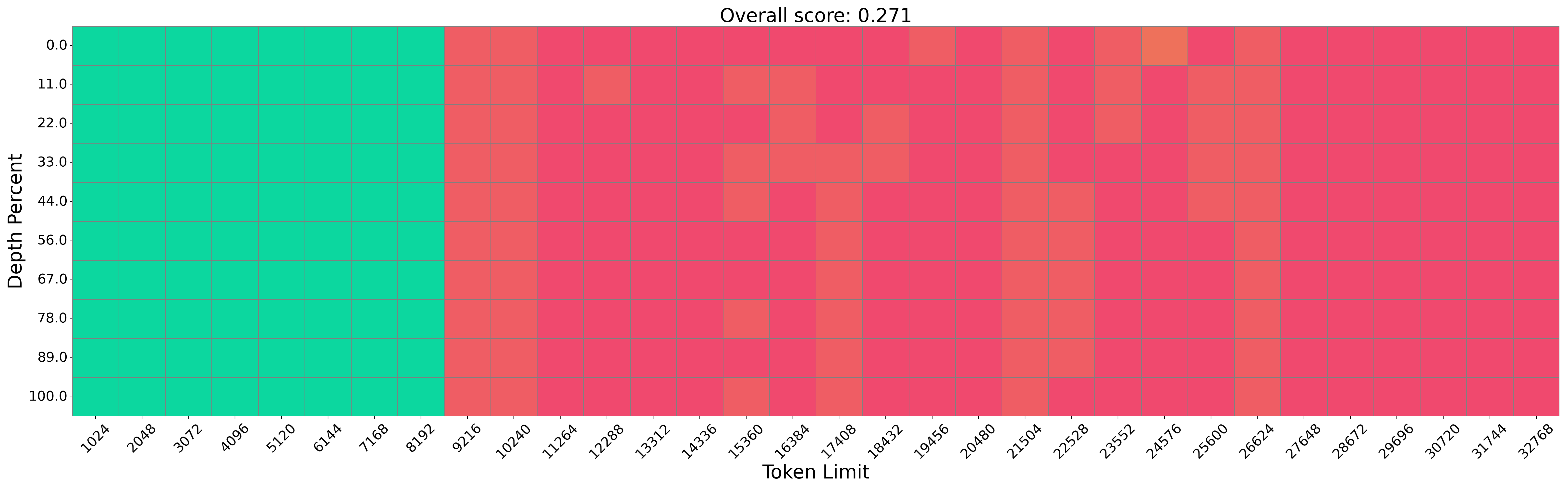}}
  
  \subfloat[StreamingLLM\label{fig:nih_llama_stream} ]{\includegraphics[width=2\columnwidth, height=0.2\linewidth]{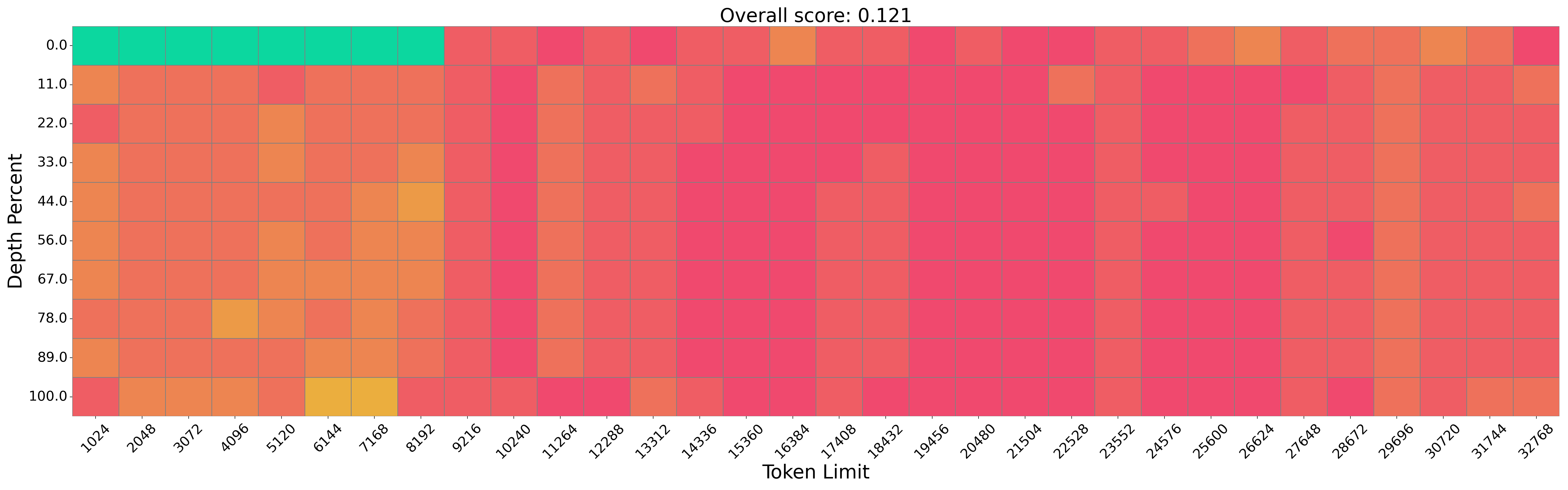}}
  
  \subfloat[H2O\label{fig:nih_llama_h2o}]{\includegraphics[width=2\columnwidth, height=0.2\linewidth]{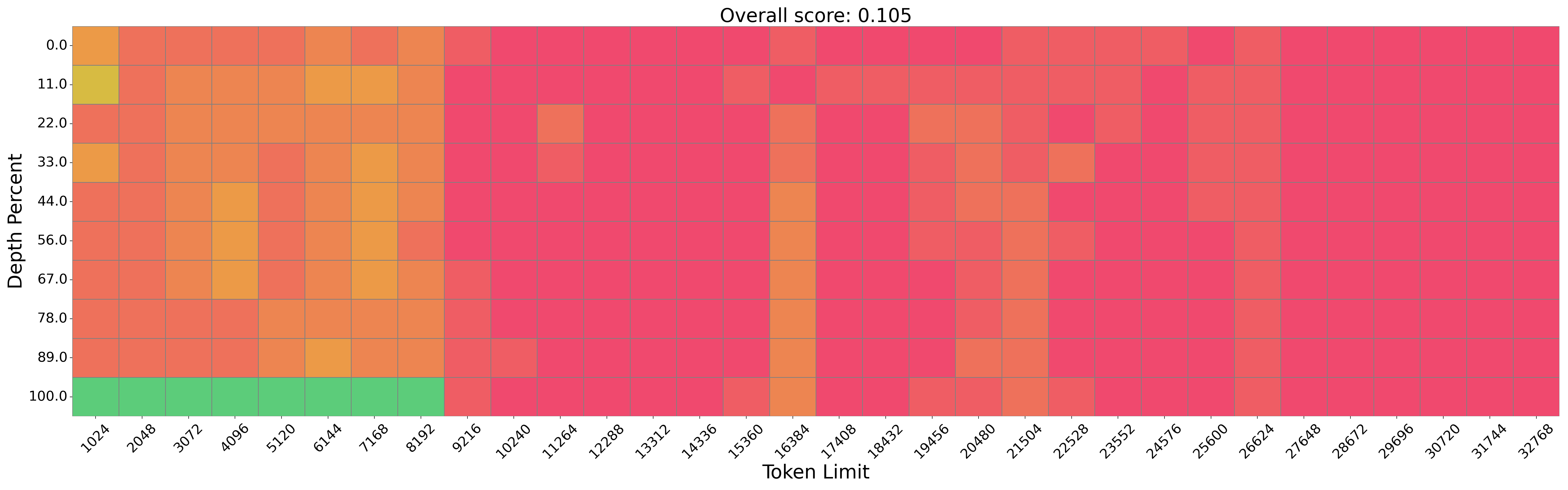}}
  
  \subfloat[SnapKV\label{fig:nih_llama_snapkv}]{\includegraphics[width=2.1\columnwidth, height=0.2\linewidth]{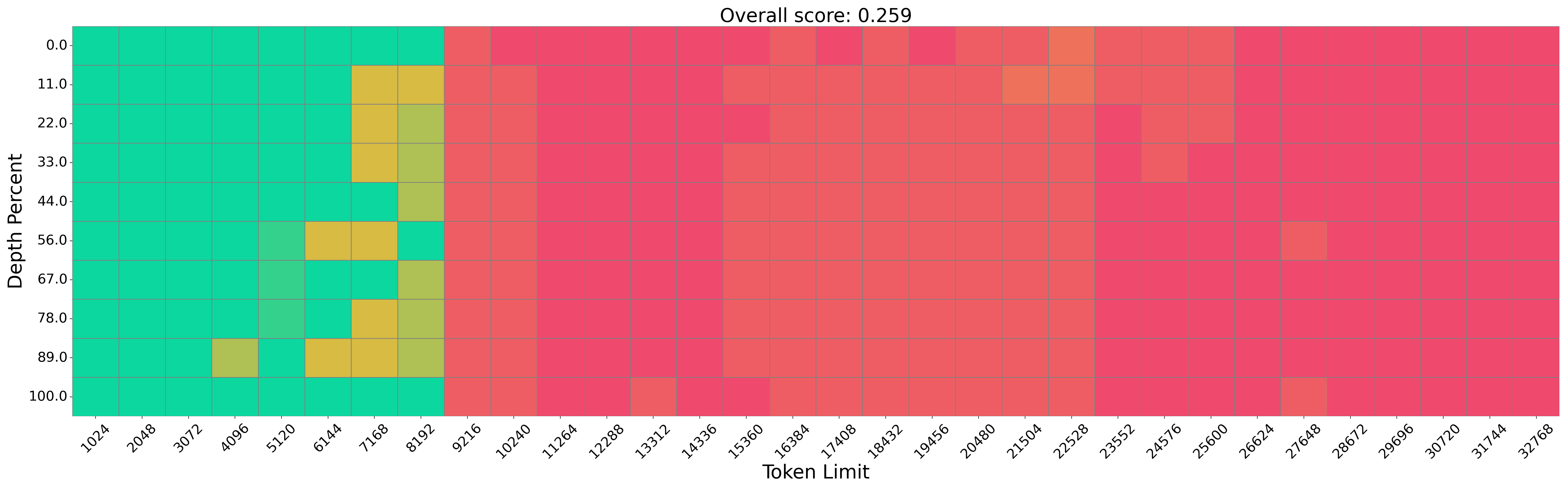}}
  
  \subfloat[PyramidKV\label{fig:llama_pyramidkv}]{\includegraphics[width=2.1\columnwidth, height=0.2\linewidth]{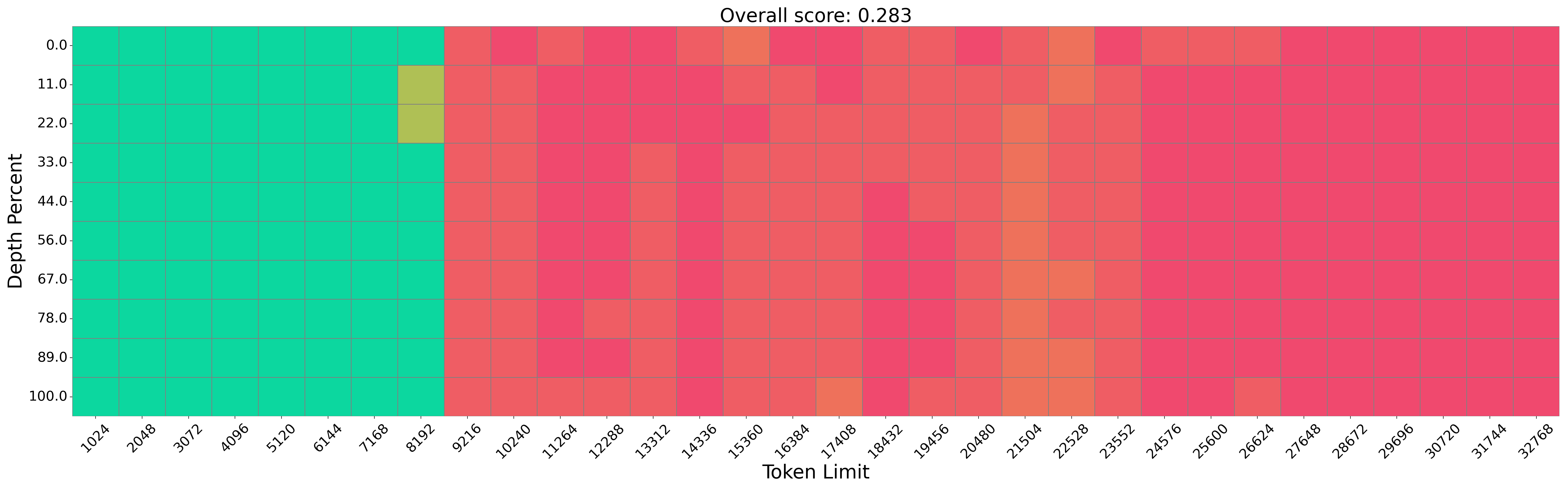}}

  \subfloat[InfiniRetri\label{fig:llama_infiniretri}]{\includegraphics[width=2.1\columnwidth, height=0.2\linewidth]{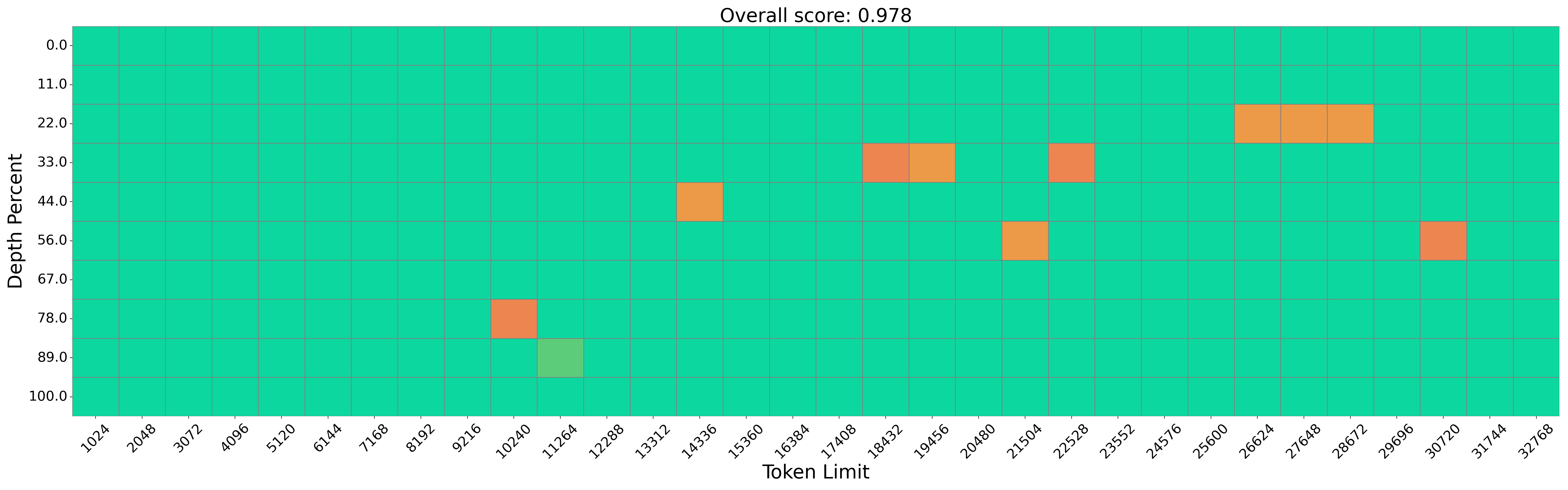}}

  \caption{\label{fig:comparion_nih_llama}
    Performance Comparison on the Needle in a Haystack Task Using Llama3-8B-Instruct.
  }
\end{figure*}

\begin{figure*}[htbp]
  \centering
  \subfloat[FullKV\label{fig:nih_mistral_full}]{\includegraphics[width=2\columnwidth, height=0.2\linewidth]{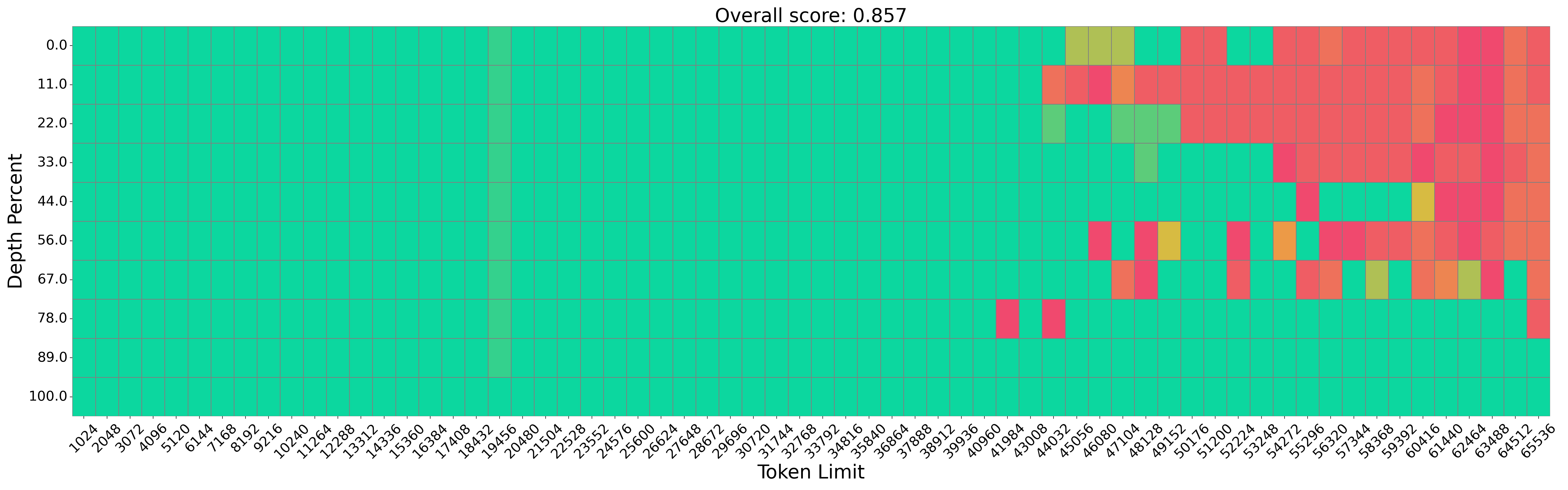}}
  
  \subfloat[InfiniRetri \label{fig:nih_mistral_infiniretri} ]{\includegraphics[width=2\columnwidth, height=0.2\linewidth]{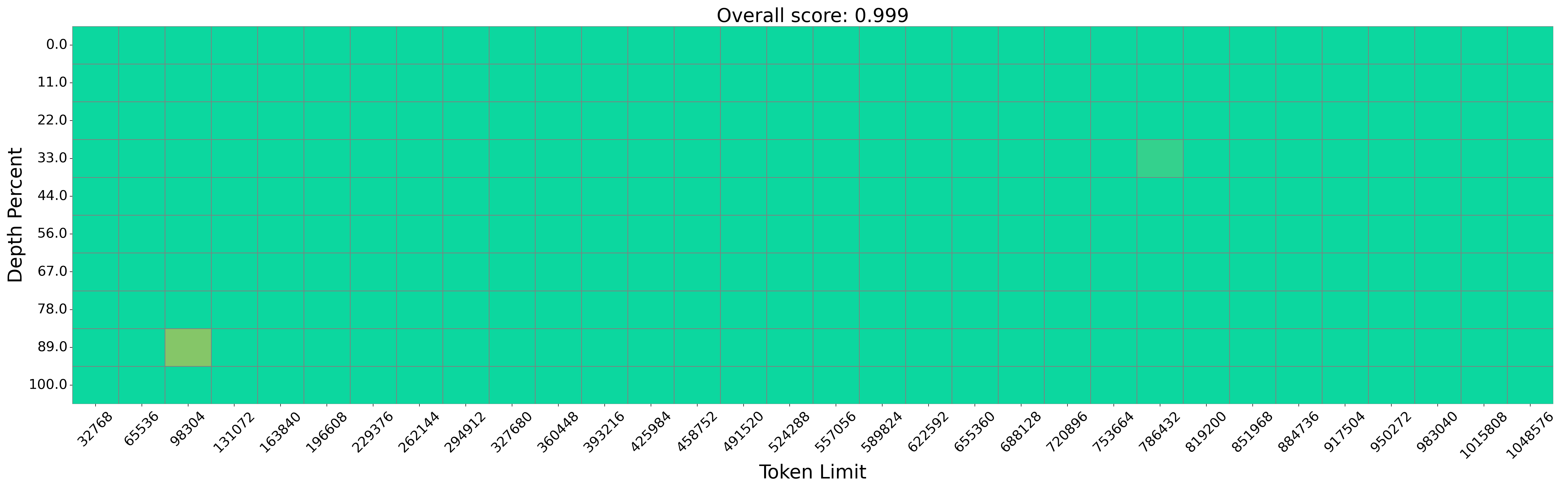}}
  \caption{ \label{fig:comparion_nih_mistral}
    Performance Comparison on the Needle in a Haystack Task Using Mistral-7B-Instructv0.2
  }
\end{figure*}

The Needle-in-a-Haystack (NIH) task demands that models accurately retrieve a specific sentence ("Needle") that can be randomly inserted into any position within a extensive document("Haystack"). This task can be visually analyzed through heatmap visualization experiments, where green indicates perfect retrieval accuracy, while other colors signify retrieval errors. This visualization provides an intuitive representation of the upper limit of LLMs’ capabilities in processing long contexts, making it widely used in evaluations. As depicted in Figure~\ref{fig:comparion_nih_llama}, we first evaluated the NIH task using the Llama3-8B-Instruct model in conjunction with FullKV, StreamingLLM, H2O, SnapKV, PyramidKV, and our InfiniRetri to assess their performance on long text inputs up to a maximum length of 32k tokens. This experimental results indicates that, while prior KV cache compression methods have shown gradual improvements, none have surpassed the performance of FullKV. In contrast, our proposed method achieves superior performance compared to FullKV, thereby surpassing the original context window (8k) processing capabilities of Llama3-8B-Instruct in the NIH task.

To further assess the effectiveness of our method on the NIH task, we extended the input length on the Mistral-7B-Instruct, which claims a context window of up to 32K tokens. We compared the results using both FullKV and our InfiniRetri, as illustrated in Figure~\ref{fig:comparion_nih_mistral}. Interestingly, our method, with identical parameter settings, outperformed Llama3 on Mistral, achieving a \textbf{100\% accuracy} rate on the NIH task. Specifically, while Mistral's NIH test could originally handle lengths up to approximately 43k tokens (in Figure~\ref{fig:nih_mistral_full}), our method enabled it to process inputs of up to \textbf{1M tokens}, without compromising accuracy (in Figure~\ref{fig:nih_mistral_infiniretri}). We further observed that, as long as an LLM \textbf{possesses sufficient retrieval ability within a limited context window, it can be empowered by our method to handle retrieval tasks of effectively unlimited length}. Building on this insight, we conducted additional experiments using the smaller, open-source model. As expected, our method expanded its effective context length from 32K to over 1M tokens, thereby enabling it to handle NIH tasks of infinitely length (as shown in Figure~\ref{fig:Qwen0.5B-compare}).

\begin{table*}[htbp]
\centering
\resizebox{\textwidth}{!}{%
\begin{tabular}{@{}l|lllllllllllll@{}}
\toprule
 &  &  & \multicolumn{3}{l}{Single-Doucument QA} & \multicolumn{3}{l}{Multi-Document QA} & \multicolumn{3}{l}{Summarization} & Avg & QA Avg \\ \cmidrule(lr){4-12}
\multirow{-2}{*}{Model} & \multirow{-2}{*}{Size} & \multirow{-2}{*}{Method} & NrtvQA & Qasper & MF-en & HotpotQA & 2WikiMQA & Musique & GovReport & QMSum & MultiNews & - & - \\ \midrule
 &  & StreamingLLM & 19.03 & 12.78 & 28.67 & 37.83 & 29.97 & 16.55 & 20.30 & 20.94 & 24.56 & 23.40 & 24.13 \\
 &  & H2O & 22.84 & 16.80 & 32.36 & 41.43 & 34.07 & 19.30 & 22.28 & 22.81 & 23.69 & 26.17 & 27.80 \\
 &  & SnapKV & 24.62 & 22.78 & 37.88 & 42.96 & 34.82 & 20.65 & 22.63 & 22.54 & 23.93 & 28.09 & 30.61 \\
 &  & PyramidKV & 24.48 & 23.51 & 36.14 & 42.33 & 31.95 & 20.73 & 23.37 & 23.01 & 24.37 & 27.76 & 29.85 \\
 & \multirow{-5}{*}{512} & Dyanamic & 24.78 & 24.76 & 36.84 & 44.13 & 33.25 & 20.82 & 23.00 & 22.76 & 24.14 & 28.27 & 30.76 \\ \cmidrule(l){2-14} 
 & - & FullKV & \textbf{25.16} & 31.81 & 39.59 & 43.09 & \textbf{36.15} & 21.77 & \textbf{28.62} & \textbf{23.34} & \textbf{26.33} & 30.65 & 32.92 \\
\multirow{-7}{*}{LlaMA3-8B-Instruct} & - & \textbf{ours} & 18.88 & \textbf{36.45} & \textbf{44.72} & \textbf{50.1} & 29.98 & \textbf{27.26} & 21.94 & 20.17 & 24.14 & \textbf{30.40} & \textbf{34.56} \\ \midrule
 &  & StreamingLLM & 20.47 & 26.97 & 32.64 & 14.31 & 14.39 & 6.82 & 25.7 & 19.31 & 24.88 & 20.61 & 19.26 \\
 &  & H2O & 22.88 & 34.28 & 41.4 & 13.3 & 14.6 & 8.31 & 23.69 & 22.07 & 22.72 & 22.58 & 22.46 \\
 &  & SnapKV & 23.86 & 38.61 & 44.65 & 15.6 & 14.62 & 9.13 & 24.56 & 22.39 & 23.07 & 24.05 & 24.41 \\
 &  & PyramidKV & 24.47 & 37.6 & 43.51 & 14.48 & 12.83 & 8.99 & 23.59 & 22.3 & 22.41 & 23.35 & 23.64 \\
 & \multirow{-5}{*}{512} & Dyanamic & 24.66 & 40.44 & 45.3 & 15.42 & 13.89 & 8.46 & 25.51 & 22.77 & 22.92 & 24.37 & 24.69 \\ \cmidrule(l){2-14} 
 & - & FullKV & 25.14 & 42.35 & 45.04 & 14.8 & 14.13 & 9.23 & 36.35 & 23.79 & 26.51 & 26.37 & 25.11 \\
\multirow{-7}{*}{Qwen2-7B-Instruct} & - & \textbf{ours} & \textbf{25.48} & 42.12 & \textbf{50.92} & \textbf{57.52} & \textbf{50.26} & \textbf{30.62} & 19.26 & 20.68 & 20.6 & \textbf{35.27} & \textbf{42.82} \\ \midrule
 &  & StreamlingLLM & 16.91 & 21.51 & 24.85 & 34.14 & 26.99 & 16.64 & 15.67 & 18.61 & 14.4 & 21.08 & 23.50 \\
 &  & H2O & 21.25 & 26.66 & 35.13 & 38.82 & 29.8 & 18.88 & 21 & 19.5 & 18.63 & 25.51 & 28.42 \\
 &  & TOVA & 22.47 & 24.26 & 37.22 & 42.26 & 28.85 & 19.97 & 19.4 & 18.7 & 17.86 & 25.66 & 29.17 \\
 &  & SnapKV & 21.02 & 27.26 & 41.25 & 45.15 & 29.23 & 22.75 & 20.47 & 20.17 & 17.75 & 27.22 & 31.11 \\
 &  & PyramidKV & 21.73 & 26.6 & 41.46 & 43.2 & 29.32 & 21.47 & 20.23 & 19.82 & 17.46 & 26.81 & 30.63 \\
 & \multirow{-6}{*}{128} & CAKE & 22.31 & 29.15 & 43.51 & 44.51 & 30.36 & 22.85 & 21.56 & 20.47 & 18.96 & 28.18 & 32.11 \\ \cmidrule(l){2-14} 
 &  & StreamlingLLM & 20.96 & 28.05 & 30.03 & 37.06 & 27.56 & 16.03 & 24.03 & 19.07 & 22.79 & 25.06 & 26.61 \\
 &  & H2O &  23.78 & 31.63 & 41.31 & 43.24 & 31.07 & 20.43 & 26.74 & 20.41 & 23.93 & 29.17 & 31.91 \\
 &  & TOVA & 26.97 & 34.51 & 45.58 & 44.32 & 32.58 & 22.83 & 26.91 & 20.75 & 23.49 & 30.08 & 34.46 \\
 &  & SnapKV & 26.63 & 35.78 & 48.11 & 45.75 & 32.2 & 23.37 & 26.71 & 21.84 & 23.18 & 31.50 & 35.30 \\
 &  & PyramidKV &  25.51 & 36.02 & 47.72 & 44.74 & 33.16 & 23.91 & 26.55 & 21.83 & 23.27 & 31.41 & 35.17 \\
 & \multirow{-6}{*}{1024} & CAKE &  \textbf{26.09} & 36.34 & 48.11 & 45.97 & 32.39 & 23.49 & \textbf{27.56} & 21.45 & 24.03 & 32.41 & 37.26 \\ \cmidrule(l){2-14} 
 & - & FullKV & 20.68 & 20.19 & 36.69 & 27.83 & 31.45 & 8.21 & 27.09 & 20.71 & \textbf{26.24} & 24.34 & 24.17 \\
\multirow{-14}{*}{\begin{tabular}[c]{@{}l@{}}Mistral-7B\\ -Instruct-v0.3\end{tabular}} & - & \textbf{ours} & 20.13 & \textbf{37.08} & \textbf{55.03} & \textbf{50.1} & \textbf{35.22} & \textbf{28.56} & 19.29 & \textbf{22.45} & 22.41 & \textbf{35.58} & \textbf{37.68} \\ \bottomrule
\end{tabular}%
}
\caption{\textbf{Performance comparison of different methods in LongBench} for full kv cache and previous kv cache compression methods: StreamingLLM, H2O, SnapKV, PyramidKV, DynamicKV, CAKE. The experimental result using LlaMA3-8B-Instruct and Qwen2-7B-Instruct in the above table are from the DynamicKV, and the using Mistral-7B-Instruct-v0.2 in the above method are from the CAKE. Bold indicates the best performance.}
\label{tab:longbench-result}
\end{table*}

\begin{table*}[htbp]
\centering
\resizebox{\textwidth}{!}{%
\begin{tabular}{@{}l|llllll@{}}
\toprule
Model & Overall(\%) & Easy(\%) & Hard(\%) & Short(\%) & Medium(\%) & Long(\%) \\ \midrule
DeepSeek-R1 & 58.3 & 66.1 & 53.4 & 62.2 & 54.4 & 59.3 \\
O1-preview & 57.7 & 66.8 & 51.1 & 62.6 & 53.5 & 58.1 \\
Gemini-2.0-Flash-Thinking & 56.0 & 62.8 & 51.9 & 61.1 & 55.2 & 49.1 \\
Qwen2.5-72B & 42.1 & 42.7 & 41.8 & 45.6 & 38.1 & 44.4 \\
\textbf{Qwen2.5-7B(InfiniRetri)} & 41.9 & \textbf{43.8} & 40.0 & 33.3 & \textbf{41.2} & \textbf{60} \\
Claude3.5 Sonnet & 41.0 & 46.9 & 37.3 & 46.1 & 38.6 & 37.0 \\
O1-mini & 37.8 & 38.9 & 37.1 & 48.6 & 33.3 & 28.6 \\
Mistral Large 24.11 & 34.4 & 38.0 & 32.2 & 41.7 & 30.7 & 29.6 \\
Llama 3.1 70B & 31.6 & 32.3 & 31.2 & 41.1 & 27.4 & 24.1 \\
\textbf{Qwen2.5-7B(origin)} & 30.0 & 30.7 & 29.6 & 40.6 & 24.02 & 24.1 \\
Llama 3.1 8B & 30.0 & 30.7 & 29.6 & 35.0 & 27.9 & 25.9 \\
Llama 3.3 70B & 29.8 & 34.4 & 27.0 & 36.7 & 27.0 & 24.1 \\ \bottomrule
\end{tabular}%
}
\caption{Performance comparison of using our method in LongBenchV2(No CoT) }
\label{tab:LongBenchV2}
\end{table*}

\subsection{Experiment on LongBench}
\label{sec:experiment_LongBench}

As illustrated in Table~\ref{tab:longbench-result}, from the overall experimental results, our method is the only one that outperforms the FullKV method across all models, with the most significant improvements observed in document QA tasks. Specifically, the performance of LlaMA3-8B-Instruct, Qwen2-7B-Instruct and Mistral-7B-Instruct-v0.3 increased by 4.9\% (32.92->34.56), 70.5\% (25.11->42.82), and 55.8\% (24.17->37.68), respectively. Among them, the Qwen2-7B-Instruct model achieved the most substantial improvement on the HotpotQA task, with a maximum increase of \textbf{288\% (14.8 -> 57.52)}. Notably, the Qwen2-7B-Instruct model's score on HotpotQA surpassed those of other models with similar parameter sizes, indicating its superior capability in short-text reasoning. This suggests that Qwen2-7B-Instruct can effectively enhance its long-document reasoning ability through our method. Similarly, the Mistral-7B-Instructv0.2, which also excels in short-text reasoning, demonstrated notable improvements in long-document QA tasks when applying our method. However, there are exceptions; for instance, LlaMA3-8B-Instruct showed minimal improvement when using our method, likely due to its inherently stable performance across varying lengths.

Subsequently, while our method achieved significant improvements in long-document QA tasks, its performance on document summarization tasks was comparatively less effective. This discrepancy may stem from the nature of summarization tasks, which typically require richer contextual information to generate high-quality outputs. Our method cannot access all relevant information at once, which limits its effectiveness in such tasks. In constast to QA and retrieval tasks, where the answer often relies on a small subset of the long context, summarization task depend heavily on a comprehensive understanding of the entire context. As such, our approach may require further optimization and refinement to better address these summarization tasks. 

To further evaluate the effectiveness of our approach, we conducted additional experiments using the latest Qwen2.5-7B-Instruct model on LongBenchV2. As results are presented in Table~\ref{tab:LongBenchV2}, after applying our method InfiniRetri, Qwen2.5-7B demonstrated a substantial improvement in handling Long and Medium length texts on LongBenchV2, bringing its overall performance on par with that of its 72B counterpart model. This result further validates that \textbf{as long as a LLMs excels in short-context scenarios, our method can effectively enhance its capability to process longer contextual texts}.

\section{Ablation Study}
\label{experiment:ablation_study}

\begin{table}[htbp]
\centering
\resizebox{0.5\textwidth}{!}{%
\begin{tabular}{@{}llll@{}}
\toprule
Cache Approach & HotpotQA & 2WikiMQA & Musique \\ \midrule
Past Key-Value State & 42.15 & 30.71 & 19.22 \\
Token IDs(Ours) & \textbf{55.95} & \textbf{49.89} & \textbf{34.6} \\ \bottomrule
\end{tabular}%
}
\caption{Compare the effectiveness of two diffierent cache approach in our method by using Qwen2.5-7B-Instruct}
\label{tab:compare_cache_method}
\end{table}

As discussed in Section~\ref{sec:4.3}, our method departs from prior KV cache methods by storing \textbf{token IDs} of the text outside the model rather than the past key-value states of each layer. Initially, it was thought that storing past key-value states offered some minor advantages in inference cost compared to store token Ids. To validate this design choice, we conducted ablation studies in our method comparing the two approaches in terms of their ability to enhance long-text processing capabilities without additional training. The experiments were performed using the Multi-Document QA dataset:  HotptoQa, 2WikiMQA and Musique in LongBenchV1 by Qwen2.7-7B-Instruct model, with results presented in Table~\ref{tab:compare_cache_method}. Our findings indicate that cache past key-value states is significantly less effective than our approach of caching token IDs. This highlights the considerable challenge of achieving effective compression in KV caches without training.

\begin{table*}[!htbp]
\centering
\resizebox{\textwidth}{!}{%
\begin{tabular}{@{}llllllllll@{}}
\toprule
\multirow{2}{*}{Model \& Method} & \multicolumn{3}{l}{Single-Document QA} & \multicolumn{3}{l}{Multi-Document QA} & \multicolumn{3}{l}{Summarization} \\ \cmidrule(l){2-10} 
 & NrtvQA & Qasper & MF-en & HotpotQA & 2wikiMQA & Musique & GovReport & QMSum & MultiNews \\ \midrule
Origin(FullKV) Average Length & 18409 & 3619 & 4559 & 9151 & 4887 & 11214 & 8734 & 10614 & 2113 \\
\begin{tabular}[c]{@{}l@{}}Llama3-8B-Instruct(InfiniRetri)\end{tabular} & 789 & 1006 & 853 & 905 & 952 & 953 & 821 & 764 & 758 \\
\begin{tabular}[c]{@{}l@{}}Mistral-7B-InstructV.02(InfiniRetri)\end{tabular} & 929 & 1049 & 974 & 987 & 1083 & 1076 & 1057 & 806 & 808 \\
\begin{tabular}[c]{@{}l@{}}Qwen2-7B-Instruct(InfiniRetri)\end{tabular} & 785 & 904 & 813 & 795 & 886 & 848 & 778 & 703 & 634 \\ \midrule
InfiniRetri Average Length & 834 & 986 & 880 & 896 & 974 & 959 & 885 & 758 & 733 \\ \bottomrule
\end{tabular}%
}
\caption{Performance comparison the effectiveness of our method on reducing the context length of LongBench task using different models.}
\label{tab:reduce_context_length}
\end{table*}

\newpage
\section{Reduced Latency \& Overhead}
\label{experiment:reduce}

As introduced above, our method employs a segment and slide window mechanism coupled with iterative processing to confine the inference length of LLMs within the range of the method parameter, while retaining only the most relevant tokens in the cache. This operational mechanism ensures that only a small fraction of the original lengthy context is actually fed into the LLMs, thereby significantly reducing inference latency and computational overhead during long-text processing. As illustrated in Table~\ref{tab:reduce_context_length}, without finely tuning method parameters, our method achieves substantial reductions in inference costs for LlaMA3-8B-Instruct, Mistral-7B-InstructV0.2, and Qwen2-7B-Instruct on Document QA tasks in LongBench. Specifically, our method retains only 4.5\% of the original input text on average on the NtvQA task (18409->834). More notably, on the HotpotQA task, it retains only 8.7\% (9152->795) of the text for Qwen2-7B-Instruct while achieving a 288\% performance improvement. These results further demonstrate that our method effectively \textbf{enhances the long-text processing capabilities of LLMs by strengthening their abilities within smaller context windows}. This observation suggests that enhancing LLMs' ability to process long texts is not only possible by scaling up the context window, but also by improving the model's capabilities within smaller windows, complemented by our method's mechanism.

\newpage
\section{Conclusion}
\label{sec:conclusion}
In this study, we innovatively proposed the concept of \textbf{attention allocation pattern alignment with retrieval-augmented} by analyzing the distribution of attention scores across each layer during LLM inference. Based on this insight, we designed a method, InfiniRetri, which can be applied to any Transformer-based model without additional training, enabling retrieval over texts of unlimited length. Unlike RAG, our method innovatively utilized the model's own attention information for accurate retrieval instead of relying on external embedding models. Compared to prior KV cache compression methods, our approach not only significantly reduced inference latency and computational overhead but also outperformed Full KV, leading to substantial improvements in realistic retrieval and long document Question-Answer (QA) tasks, which demonstrates substantial practical value in scenarios involving extremely long contexts. 

Notably, our method, for the first time, successfully addressed the Needle-In-a-Haystack (NIH) task: if an LLM could accurately retrieve answers within a limited context window, our method enabled it correct retrieval from texts of infinitely length. Building on this achievement, our method offers an alternative perspective for research focused solely on extending context windows: Enhancing the model's internal capabilities within a smaller context window and integrating our method's mechanism were able to achieve better long-context performance. Compared to retrieval and QA tasks, our method underperformed in long document summarization tasks. These limitations also point to directions for future research and improvement. Overall, our work provides a novel and effective solution for long-text processing in LLMs, paving the way for future research on efficient retrieval and context extension.

\newpage
\bibliography{custom}

\begin{thebibliography}{54}
\providecommand{\natexlab}[1]{#1}

\bibitem[{Achiam et~al.(2023)Achiam, Adler, Agarwal, Ahmad, Akkaya, Aleman, Almeida, Altenschmidt, Altman, Anadkat et~al.}]{achiam2023gpt}
Josh Achiam, Steven Adler, Sandhini Agarwal, Lama Ahmad, Ilge Akkaya, Florencia~Leoni Aleman, Diogo Almeida, Janko Altenschmidt, Sam Altman, Shyamal Anadkat, et~al. 2023.
\newblock Gpt-4 technical report.
\newblock \emph{arXiv preprint arXiv:2303.08774}.

\bibitem[{Bai et~al.(2023)Bai, Lv, Zhang, Lyu, Tang, Huang, Du, Liu, Zeng, Hou et~al.}]{bai2023longbench}
Yushi Bai, Xin Lv, Jiajie Zhang, Hongchang Lyu, Jiankai Tang, Zhidian Huang, Zhengxiao Du, Xiao Liu, Aohan Zeng, Lei Hou, et~al. 2023.
\newblock Longbench: A bilingual, multitask benchmark for long context understanding.
\newblock \emph{arXiv preprint arXiv:2308.14508}.

\bibitem[{Bai et~al.(2024)Bai, Tu, Zhang, Peng, Wang, Lv, Cao, Xu, Hou, Dong et~al.}]{bai2024longbench}
Yushi Bai, Shangqing Tu, Jiajie Zhang, Hao Peng, Xiaozhi Wang, Xin Lv, Shulin Cao, Jiazheng Xu, Lei Hou, Yuxiao Dong, et~al. 2024.
\newblock Longbench v2: Towards deeper understanding and reasoning on realistic long-context multitasks.
\newblock \emph{arXiv preprint arXiv:2412.15204}.

\bibitem[{Cai et~al.(2024{\natexlab{a}})Cai, Zhang, Gao, Liu, Liu, Lu, Xiong, Dong, Chang, Hu et~al.}]{cai2024pyramidkv}
Zefan Cai, Yichi Zhang, Bofei Gao, Yuliang Liu, Tianyu Liu, Keming Lu, Wayne Xiong, Yue Dong, Baobao Chang, Junjie Hu, et~al. 2024{\natexlab{a}}.
\newblock Pyramidkv: Dynamic kv cache compression based on pyramidal information funneling.
\newblock \emph{arXiv preprint arXiv:2406.02069}.

\bibitem[{Cai et~al.(2024{\natexlab{b}})Cai, Cao, Chen, Chen, Chen, Chen, Chen, Chen, Chen, Chu et~al.}]{cai2024internlm2}
Zheng Cai, Maosong Cao, Haojiong Chen, Kai Chen, Keyu Chen, Xin Chen, Xun Chen, Zehui Chen, Zhi Chen, Pei Chu, et~al. 2024{\natexlab{b}}.
\newblock Internlm2 technical report.
\newblock \emph{arXiv preprint arXiv:2403.17297}.

\bibitem[{Chen et~al.(2023{\natexlab{a}})Chen, Wong, Chen, and Tian}]{chen2023extending}
Shouyuan Chen, Sherman Wong, Liangjian Chen, and Yuandong Tian. 2023{\natexlab{a}}.
\newblock Extending context window of large language models via positional interpolation.
\newblock \emph{arXiv preprint arXiv:2306.15595}.

\bibitem[{Chen et~al.(2023{\natexlab{b}})Chen, Qian, Tang, Lai, Liu, Han, and Jia}]{chen2023longlora}
Yukang Chen, Shengju Qian, Haotian Tang, Xin Lai, Zhijian Liu, Song Han, and Jiaya Jia. 2023{\natexlab{b}}.
\newblock Longlora: Efficient fine-tuning of long-context large language models.
\newblock \emph{arXiv preprint arXiv:2309.12307}.

\bibitem[{Chiang et~al.(2023)Chiang, Li, Lin, Sheng, Wu, Zhang, Zheng, Zhuang, Zhuang, Gonzalez et~al.}]{chiang2023vicuna}
Wei-Lin Chiang, Zhuohan Li, Zi~Lin, Ying Sheng, Zhanghao Wu, Hao Zhang, Lianmin Zheng, Siyuan Zhuang, Yonghao Zhuang, Joseph~E Gonzalez, et~al. 2023.
\newblock Vicuna: An open-source chatbot impressing gpt-4 with 90\%* chatgpt quality.
\newblock \emph{See https://vicuna. lmsys. org (accessed 14 April 2023)}, 2(3):6.

\bibitem[{Cohere(2024)}]{command}
Cohere. 2024.
\newblock Command r: Retrieval-augmented generation at production scale.
\newblock \emph{https://txt.cohere.com/command-r.}

\bibitem[{Dasigi et~al.(2021)Dasigi, Lo, Beltagy, Cohan, Smith, and Gardner}]{Dasigi_Lo_Beltagy_Cohan_Smith_Gardner_2021}
Pradeep Dasigi, Kyle Lo, Iz~Beltagy, Arman Cohan, Noah~A. Smith, and Matt Gardner. 2021.
\newblock \href {https://doi.org/10.18653/v1/2021.naacl-main.365} {A dataset of information-seeking questions and answers anchored in research papers}.
\newblock In \emph{Proceedings of the 2021 Conference of the North American Chapter of the Association for Computational Linguistics: Human Language Technologies}.

\bibitem[{Dubey et~al.(2024)Dubey, Jauhri, Pandey, Kadian, Al-Dahle, Letman, Mathur, Schelten, Yang, Fan et~al.}]{dubey2024llama}
Abhimanyu Dubey, Abhinav Jauhri, Abhinav Pandey, Abhishek Kadian, Ahmad Al-Dahle, Aiesha Letman, Akhil Mathur, Alan Schelten, Amy Yang, Angela Fan, et~al. 2024.
\newblock The llama 3 herd of models.
\newblock \emph{arXiv preprint arXiv:2407.21783}.

\bibitem[{Fabbri et~al.(2019{\natexlab{a}})Fabbri, Li, She, Li, and Radev}]{fabbri2019multi}
Alexander~R Fabbri, Irene Li, Tianwei She, Suyi Li, and Dragomir~R Radev. 2019{\natexlab{a}}.
\newblock Multi-news: A large-scale multi-document summarization dataset and abstractive hierarchical model.
\newblock \emph{arXiv preprint arXiv:1906.01749}.

\bibitem[{Fabbri et~al.(2019{\natexlab{b}})Fabbri, Li, She, Li, and Radev}]{fabbri2019multi_news}
Alexander~R Fabbri, Irene Li, Tianwei She, Suyi Li, and Dragomir~R Radev. 2019{\natexlab{b}}.
\newblock Multi-news: A large-scale multi-document summarization dataset and abstractive hierarchical model.
\newblock \emph{arXiv preprint arXiv:1906.01749}.

\bibitem[{Fu et~al.(2024)Fu, Panda, Niu, Yue, Hajishirzi, Kim, and Peng}]{fu2024data}
Yao Fu, Rameswar Panda, Xinyao Niu, Xiang Yue, Hannaneh Hajishirzi, Yoon Kim, and Hao Peng. 2024.
\newblock Data engineering for scaling language models to 128k context.
\newblock \emph{arXiv preprint arXiv:2402.10171}.

\bibitem[{Gao et~al.(2024)Gao, Wettig, Yen, and Chen}]{gao2024train}
Tianyu Gao, Alexander Wettig, Howard Yen, and Danqi Chen. 2024.
\newblock How to train long-context language models (effectively).
\newblock \emph{arXiv preprint arXiv:2410.02660}.

\bibitem[{Gao et~al.(2023)Gao, Xiong, Gao, Jia, Pan, Bi, Dai, Sun, and Wang}]{gao2023retrieval}
Yunfan Gao, Yun Xiong, Xinyu Gao, Kangxiang Jia, Jinliu Pan, Yuxi Bi, Yi~Dai, Jiawei Sun, and Haofen Wang. 2023.
\newblock Retrieval-augmented generation for large language models: A survey.
\newblock \emph{arXiv preprint arXiv:2312.10997}.

\bibitem[{GLM et~al.(2024)GLM, Zeng, Xu, Wang, Zhang, Yin, Zhang, Rojas, Feng, Zhao et~al.}]{glm2024chatglm}
Team GLM, Aohan Zeng, Bin Xu, Bowen Wang, Chenhui Zhang, Da~Yin, Dan Zhang, Diego Rojas, Guanyu Feng, Hanlin Zhao, et~al. 2024.
\newblock Chatglm: A family of large language models from glm-130b to glm-4 all tools.
\newblock \emph{arXiv preprint arXiv:2406.12793}.

\bibitem[{He et~al.(2024)He, Karlinsky, Kim, McAuley, Krotov, and Feris}]{he2024camelot}
Zexue He, Leonid Karlinsky, Donghyun Kim, Julian McAuley, Dmitry Krotov, and Rogerio Feris. 2024.
\newblock Camelot: Towards large language models with training-free consolidated associative memory.
\newblock \emph{arXiv preprint arXiv:2402.13449}.

\bibitem[{Ho et~al.(2020)Ho, Nguyen, Sugawara, and Aizawa}]{ho2020mutltihopqa}
Xanh Ho, Anh-Khoa~Duong Nguyen, Saku Sugawara, and Akiko Aizawa. 2020.
\newblock Constructing a multi-hop qa dataset for comprehensive evaluation of reasoning steps.
\newblock \emph{arXiv preprint arXiv:2011.01060}.

\bibitem[{Hsieh et~al.(2024)Hsieh, Sun, Kriman, Acharya, Rekesh, Jia, Zhang, and Ginsburg}]{hsieh2024ruler}
Cheng-Ping Hsieh, Simeng Sun, Samuel Kriman, Shantanu Acharya, Dima Rekesh, Fei Jia, Yang Zhang, and Boris Ginsburg. 2024.
\newblock Ruler: What's the real context size of your long-context language models?
\newblock \emph{arXiv preprint arXiv:2404.06654}.

\bibitem[{Hu et~al.(2021)Hu, Shen, Wallis, Allen-Zhu, Li, Wang, Wang, and Chen}]{hu2021lora}
Edward~J Hu, Yelong Shen, Phillip Wallis, Zeyuan Allen-Zhu, Yuanzhi Li, Shean Wang, Lu~Wang, and Weizhu Chen. 2021.
\newblock Lora: Low-rank adaptation of large language models.
\newblock \emph{arXiv preprint arXiv:2106.09685}.

\bibitem[{Huang et~al.(2021)Huang, Cao, Parulian, Ji, and Wang}]{huang2021government_report}
Luyang Huang, Shuyang Cao, Nikolaus Parulian, Heng Ji, and Lu~Wang. 2021.
\newblock Efficient attentions for long document summarization.
\newblock \emph{arXiv preprint arXiv:2104.02112}.

\bibitem[{Hwang et~al.(2024)Hwang, Wang, Huo, Sim, and Mengibar}]{hwang2024transformerfam}
Dongseong Hwang, Weiran Wang, Zhuoyuan Huo, Khe~Chai Sim, and Pedro~Moreno Mengibar. 2024.
\newblock Transformerfam: Feedback attention is working memory.
\newblock \emph{arXiv preprint arXiv:2404.09173}.

\bibitem[{Jiang et~al.(2023)Jiang, Sablayrolles, Mensch, Bamford, Chaplot, Casas, Bressand, Lengyel, Lample, Saulnier et~al.}]{jiang2023mistral}
Albert~Q Jiang, Alexandre Sablayrolles, Arthur Mensch, Chris Bamford, Devendra~Singh Chaplot, Diego de~las Casas, Florian Bressand, Gianna Lengyel, Guillaume Lample, Lucile Saulnier, et~al. 2023.
\newblock Mistral 7b.
\newblock \emph{arXiv preprint arXiv:2310.06825}.

\bibitem[{Jin et~al.(2024)Jin, Han, Yang, Jiang, Liu, Chang, Chen, and Hu}]{jin2024llm}
Hongye Jin, Xiaotian Han, Jingfeng Yang, Zhimeng Jiang, Zirui Liu, Chia-Yuan Chang, Huiyuan Chen, and Xia Hu. 2024.
\newblock Llm maybe longlm: Self-extend llm context window without tuning.
\newblock \emph{arXiv preprint arXiv:2401.01325}.

\bibitem[{Kim et~al.(2024)Kim, Shim, Choi, and Chang}]{kim2024infinipot}
Minsoo Kim, Kyuhong Shim, Jungwook Choi, and Simyung Chang. 2024.
\newblock Infinipot: Infinite context processing on memory-constrained llms.
\newblock \emph{arXiv preprint arXiv:2410.01518}.

\bibitem[{Li et~al.(2024)Li, Huang, Yang, Venkitesh, Locatelli, Ye, Cai, Lewis, and Chen}]{li2024snapkv}
Yuhong Li, Yingbing Huang, Bowen Yang, Bharat Venkitesh, Acyr Locatelli, Hanchen Ye, Tianle Cai, Patrick Lewis, and Deming Chen. 2024.
\newblock Snapkv: Llm knows what you are looking for before generation.
\newblock \emph{arXiv preprint arXiv:2404.14469}.

\bibitem[{Liu et~al.(2024{\natexlab{a}})Liu, Feng, Xue, Wang, Wu, Lu, Zhao, Deng, Zhang, Ruan et~al.}]{liu2024deepseek}
Aixin Liu, Bei Feng, Bing Xue, Bingxuan Wang, Bochao Wu, Chengda Lu, Chenggang Zhao, Chengqi Deng, Chenyu Zhang, Chong Ruan, et~al. 2024{\natexlab{a}}.
\newblock Deepseek-v3 technical report.
\newblock \emph{arXiv preprint arXiv:2412.19437}.

\bibitem[{Liu et~al.(2024{\natexlab{b}})Liu, Lin, Hewitt, Paranjape, Bevilacqua, Petroni, and Liang}]{liu2024lost}
Nelson~F Liu, Kevin Lin, John Hewitt, Ashwin Paranjape, Michele Bevilacqua, Fabio Petroni, and Percy Liang. 2024{\natexlab{b}}.
\newblock Lost in the middle: How language models use long contexts.
\newblock \emph{Transactions of the Association for Computational Linguistics}, 12:157--173.

\bibitem[{Munkhdalai et~al.(2024)Munkhdalai, Faruqui, and Gopal}]{munkhdalai2024leave}
Tsendsuren Munkhdalai, Manaal Faruqui, and Siddharth Gopal. 2024.
\newblock Leave no context behind: Efficient infinite context transformers with infini-attention.
\newblock \emph{arXiv preprint arXiv:2404.07143}.

\bibitem[{OpenAI(2024)}]{openai-o1}
OpenAI. 2024.
\newblock Learning to reason with llms.
\newblock \emph{https://openai.com/index/learning-to-reason-with-llms/.}

\bibitem[{Peng et~al.(2023)Peng, Quesnelle, Fan, and Shippole}]{peng2023yarn}
Bowen Peng, Jeffrey Quesnelle, Honglu Fan, and Enrico Shippole. 2023.
\newblock Yarn: Efficient context window extension of large language models.
\newblock \emph{arXiv preprint arXiv:2309.00071}.

\bibitem[{Qin et~al.(2025)Qin, Cao, Lin, Hu, Fan, Cheng, Lin, and Li}]{qin2025cake}
Ziran Qin, Yuchen Cao, Mingbao Lin, Wen Hu, Shixuan Fan, Ke~Cheng, Weiyao Lin, and Jianguo Li. 2025.
\newblock \href {https://openreview.net/forum?id=EQgEMAD4kv} {{CAKE}: Cascading and adaptive {KV} cache eviction with layer preferences}.
\newblock In \emph{The Thirteenth International Conference on Learning Representations}.

\bibitem[{Roziere et~al.(2023)Roziere, Gehring, Gloeckle, Sootla, Gat, Tan, Adi, Liu, Sauvestre, Remez et~al.}]{roziere2023code}
Baptiste Roziere, Jonas Gehring, Fabian Gloeckle, Sten Sootla, Itai Gat, Xiaoqing~Ellen Tan, Yossi Adi, Jingyu Liu, Romain Sauvestre, Tal Remez, et~al. 2023.
\newblock Code llama: Open foundation models for code.
\newblock \emph{arXiv preprint arXiv:2308.12950}.

\bibitem[{Su et~al.(2024)Su, Ahmed, Lu, Pan, Bo, and Liu}]{su2024roformer}
Jianlin Su, Murtadha Ahmed, Yu~Lu, Shengfeng Pan, Wen Bo, and Yunfeng Liu. 2024.
\newblock Roformer: Enhanced transformer with rotary position embedding.
\newblock \emph{Neurocomputing}, 568:127063.

\bibitem[{Team et~al.(2024)Team, Georgiev, Lei, Burnell, Bai, Gulati, Tanzer, Vincent, Pan, Wang et~al.}]{team2024gemini}
Gemini Team, Petko Georgiev, Ving~Ian Lei, Ryan Burnell, Libin Bai, Anmol Gulati, Garrett Tanzer, Damien Vincent, Zhufeng Pan, Shibo Wang, et~al. 2024.
\newblock Gemini 1.5: Unlocking multimodal understanding across millions of tokens of context.
\newblock \emph{arXiv preprint arXiv:2403.05530}.

\bibitem[{Team(2025)}]{qwen2.5-1m}
Qwen Team. 2025.
\newblock Qwen2.5-1m: Deploy your own qwen with context length up to 1m tokens.
\newblock \emph{https://qwenlm.github.io/blog/qwen2.5-1m/}.

\bibitem[{Touvron et~al.(2023)Touvron, Lavril, Izacard, Martinet, Lachaux, Lacroix, Rozi{\`e}re, Goyal, Hambro, Azhar et~al.}]{touvron2023llama}
Hugo Touvron, Thibaut Lavril, Gautier Izacard, Xavier Martinet, Marie-Anne Lachaux, Timoth{\'e}e Lacroix, Baptiste Rozi{\`e}re, Naman Goyal, Eric Hambro, Faisal Azhar, et~al. 2023.
\newblock Llama: Open and efficient foundation language models.
\newblock \emph{arXiv preprint arXiv:2302.13971}.

\bibitem[{Vaswani(2017)}]{vaswani2017attention}
A~Vaswani. 2017.
\newblock Attention is all you need.
\newblock \emph{Advances in Neural Information Processing Systems}.

\bibitem[{Wang et~al.(2024)Wang, Salmani, Omidi, Ren, Rezagholizadeh, and Eshaghi}]{wang2024beyond}
Xindi Wang, Mahsa Salmani, Parsa Omidi, Xiangyu Ren, Mehdi Rezagholizadeh, and Armaghan Eshaghi. 2024.
\newblock Beyond the limits: A survey of techniques to extend the context length in large language models.
\newblock \emph{arXiv preprint arXiv:2402.02244}.

\bibitem[{Wei et~al.(2022)Wei, Wang, Schuurmans, Bosma, Xia, Chi, Le, Zhou et~al.}]{wei2022chain}
Jason Wei, Xuezhi Wang, Dale Schuurmans, Maarten Bosma, Fei Xia, Ed~Chi, Quoc~V Le, Denny Zhou, et~al. 2022.
\newblock Chain-of-thought prompting elicits reasoning in large language models.
\newblock \emph{Advances in neural information processing systems}, 35:24824--24837.

\bibitem[{Xiao et~al.(2024{\natexlab{a}})Xiao, Zhang, Han, Xiao, Lin, Zhang, Liu, and Sun}]{xiao2024infllm}
Chaojun Xiao, Pengle Zhang, Xu~Han, Guangxuan Xiao, Yankai Lin, Zhengyan Zhang, Zhiyuan Liu, and Maosong Sun. 2024{\natexlab{a}}.
\newblock Infllm: Training-free long-context extrapolation for llms with an efficient context memory.
\newblock In \emph{The Thirty-eighth Annual Conference on Neural Information Processing Systems}.

\bibitem[{Xiao et~al.(2024{\natexlab{b}})Xiao, Tang, Zuo, Guo, Yang, Tang, Fu, and Han}]{xiao2024duoattention}
Guangxuan Xiao, Jiaming Tang, Jingwei Zuo, Junxian Guo, Shang Yang, Haotian Tang, Yao Fu, and Song Han. 2024{\natexlab{b}}.
\newblock Duoattention: Efficient long-context llm inference with retrieval and streaming heads.
\newblock \emph{arXiv preprint arXiv:2410.10819}.

\bibitem[{Xiao et~al.(2023)Xiao, Tian, Chen, Han, and Lewis}]{xiao2023efficient}
Guangxuan Xiao, Yuandong Tian, Beidi Chen, Song Han, and Mike Lewis. 2023.
\newblock Efficient streaming language models with attention sinks.
\newblock \emph{arXiv preprint arXiv:2309.17453}.

\bibitem[{Xiong et~al.(2023)Xiong, Liu, Molybog, Zhang, Bhargava, Hou, Martin, Rungta, Sankararaman, Oguz et~al.}]{xiong2023effective}
Wenhan Xiong, Jingyu Liu, Igor Molybog, Hejia Zhang, Prajjwal Bhargava, Rui Hou, Louis Martin, Rashi Rungta, Karthik~Abinav Sankararaman, Barlas Oguz, et~al. 2023.
\newblock Effective long-context scaling of foundation models.
\newblock \emph{arXiv preprint arXiv:2309.16039}.

\bibitem[{Yang et~al.(2024)Yang, Yang, Zhang, Hui, Zheng, Yu, Li, Liu, Huang, Wei et~al.}]{yang2024qwen2}
An~Yang, Baosong Yang, Beichen Zhang, Binyuan Hui, Bo~Zheng, Bowen Yu, Chengyuan Li, Dayiheng Liu, Fei Huang, Haoran Wei, et~al. 2024.
\newblock Qwen2. 5 technical report.
\newblock \emph{arXiv preprint arXiv:2412.15115}.

\bibitem[{Yang et~al.(2018)Yang, Qi, Zhang, Bengio, Cohen, Salakhutdinov, and Manning}]{Yang_Qi_Zhang_Bengio_Cohen_Salakhutdinov_Manning_2018}
Zhilin Yang, Peng Qi, Saizheng Zhang, Yoshua Bengio, William Cohen, Ruslan Salakhutdinov, and Christopher~D. Manning. 2018.
\newblock \href {https://doi.org/10.18653/v1/d18-1259} {Hotpotqa: A dataset for diverse, explainable multi-hop question answering}.
\newblock In \emph{Proceedings of the 2018 Conference on Empirical Methods in Natural Language Processing}.

\bibitem[{Yu et~al.(2024)Yu, Xu, and Akkiraju}]{yu2024defense}
Tan Yu, Anbang Xu, and Rama Akkiraju. 2024.
\newblock In defense of rag in the era of long-context language models.
\newblock \emph{arXiv preprint arXiv:2409.01666}.

\bibitem[{Zhang et~al.(2024)Zhang, Yu, Dong, Li, Su, Chu, and Yu}]{zhang2024mm}
Duzhen Zhang, Yahan Yu, Jiahua Dong, Chenxing Li, Dan Su, Chenhui Chu, and Dong Yu. 2024.
\newblock Mm-llms: Recent advances in multimodal large language models.
\newblock \emph{arXiv preprint arXiv:2401.13601}.

\bibitem[{Zhang et~al.(2023)Zhang, Sheng, Zhou, Chen, Zheng, Cai, Song, Tian, R{\'e}, Barrett et~al.}]{zhang2023h2o}
Zhenyu Zhang, Ying Sheng, Tianyi Zhou, Tianlong Chen, Lianmin Zheng, Ruisi Cai, Zhao Song, Yuandong Tian, Christopher R{\'e}, Clark Barrett, et~al. 2023.
\newblock H2o: Heavy-hitter oracle for efficient generative inference of large language models.
\newblock \emph{Advances in Neural Information Processing Systems}, 36:34661--34710.

\bibitem[{Zhao et~al.(2024)Zhao, Zhang, Yu, Wang, Geng, Fu, Yang, Zhang, and Cui}]{zhao2024retrieval}
Penghao Zhao, Hailin Zhang, Qinhan Yu, Zhengren Wang, Yunteng Geng, Fangcheng Fu, Ling Yang, Wentao Zhang, and Bin Cui. 2024.
\newblock Retrieval-augmented generation for ai-generated content: A survey.
\newblock \emph{arXiv preprint arXiv:2402.19473}.

\bibitem[{Zhong et~al.(2021)Zhong, Yin, Yu, Zaidi, Mutuma, Jha, Awadallah, Celikyilmaz, Liu, Qiu et~al.}]{zhong2021qmsum}
Ming Zhong, Da~Yin, Tao Yu, Ahmad Zaidi, Mutethia Mutuma, Rahul Jha, Ahmed~Hassan Awadallah, Asli Celikyilmaz, Yang Liu, Xipeng Qiu, et~al. 2021.
\newblock Qmsum: A new benchmark for query-based multi-domain meeting summarization.
\newblock \emph{arXiv preprint arXiv:2104.05938}.

\bibitem[{Zhong et~al.(2024)Zhong, Liu, Pan, Zhang, Zhou, Liang, Wu, Lyu, Shu, Yu et~al.}]{zhong2024evaluation}
Tianyang Zhong, Zhengliang Liu, Yi~Pan, Yutong Zhang, Yifan Zhou, Shizhe Liang, Zihao Wu, Yanjun Lyu, Peng Shu, Xiaowei Yu, et~al. 2024.
\newblock Evaluation of openai o1: Opportunities and challenges of agi.
\newblock \emph{arXiv preprint arXiv:2409.18486}.

\bibitem[{Zhou et~al.(2024)Zhou, Wang, Zeng, Guo, Liu, Shen, Zhang, and Ding}]{zhou2024dynamickv}
Xiabin Zhou, Wenbin Wang, Minyan Zeng, Jiaxian Guo, Xuebo Liu, Li~Shen, Min Zhang, and Liang Ding. 2024.
\newblock Dynamickv: Task-aware adaptive kv cache compression for long context llms.
\newblock \emph{arXiv preprint arXiv:2412.14838}.

\end{thebibliography}

\clearpage

\appendix
\section{Implementation details}

\label{experiment:implementation}
\noindent\textbf{Models and Infinitely Length.}
In order to compare with the latest methods, we selected five open-sourced leading LLMs from HuggingFace including Llama3-8B-instruct\footnote{https://huggingface.co/meta-llama/Meta-Llama-3-8B-Instruct}, Mistral-7B-Instruct-v0.2\footnote{https://huggingface.co/mistralai/Mistral-7B-Instruct-v0.2}, Mistral-7B-Instruct-v0.3\footnote{https://huggingface.co/mistralai/Mistral-7B-Instruct-v0.3} and Qwen2-7B-Instruct\footnote{https://huggingface.co/Qwen/Qwen2-7B-Instruct}, Qwen2.5-7B Instruct\footnote{https://huggingface.co/Qwen/Qwen2.5-7B-Instruct}. Since our method enables the processing of infinitely long texts, we do not impose restrictions on the maximum input length during the evaluation of the above models.

\noindent\textbf{Baselines.}
We compare InfiniRetri with six leading approaches in the key-value cache compression domain from various periods up to the present, as follows: (1)\textbf{StreamingLLM}:which innovatively proposed the attention sinks and used KV caches to retain the most recent token. (2)\textbf{H2O} employed a Heavy Hitter Oracle to manage KV cache.(3)\textbf{SnapKV} innovatively used attention feature to retain token in the cache. (4)\textbf{PyramidKV} introduced a pyramid pattern which innovatively design to dynamically save tokens in the cache. (5)\textbf{DynamicKV} innovatively designed a strategy to dynamically adjust cache to retain token at each layer based on the attention distribution. (6)\textbf{CAKE} proposed an adaptive cache allocation strategy on layer preferences to dynamically dajust the KV cache size of each layer. Additional, our the experimental data for these comparisons are primarily sourced from lastest work of DynamicKV and CAKE.

\noindent\textbf{Datasets.}
In Section~\ref{sec:experiment_NIH}, we conducted the "Fact Retrieval Across Context Lengths" (Needle In A Haystack) experiment using the PaulGrahamEssays dataset, following the experimental setup from PyramidKV\citep{cai2024pyramidkv}. In Section~\ref{sec:experiment_LongBench}, LongBenchV1\citep{bai2023longbench} is a comprehensive benchmark for evaluating long-text processing capabilities of LLMs, which encompassing diverse long-text datasets across various tasks types. We selected nine datasets that are currently suitable for our method, including the Single-Document QA: NarrativeQA\citep{Yang_Qi_Zhang_Bengio_Cohen_Salakhutdinov_Manning_2018}, Qasper\citep{Dasigi_Lo_Beltagy_Cohan_Smith_Gardner_2021}, MultiFieldQAen, the Multi-Document QA dataset: HotpotQA\citep{Yang_Qi_Zhang_Bengio_Cohen_Salakhutdinov_Manning_2018}, 2WikiMultihopQA\citep{ho2020mutltihopqa}, and the Summarization dataset GovReport\citep{huang2021government_report}, QMSum\citep{zhong2021qmsum}, MultiNews\citep{fabbri2019multi_news}. The datasets cover a range of real-world application scenarios. Additionally, we conducted experiments on the upgraded LongBenchV2\citep{bai2024longbench}, which challenges LLMs to answer multiple-choice questions based on fresher and significantly longer contexts, thereby avoiding potential biases from prior exposure to similar training data.

\noindent\textbf{Method Parameters Setup.} 
Our method relies heavily on three parameters: \textit{Chunk Size}, \textit{Phrase Token Num}, and \textit{TopK}. Specifically, \textit{Chunk Size} and \textit{TopK} directly influence the input length that the LLMs process during each inference, while \textit{Phrase Token Num} determines the granularity of token retrieval in our method. Given the performance variability across different models and text lengths, as well as the diverse optimal parameters for varying question types and difficulties, we opted for a unified parameter set (\textit{Chunk Size}=1024,  \textit{Topk}=300, \textit{Phrase Token Num}=15) to ensure fair and efficient evaluation across all models and tasks. Although this approach may slightly understate our method's potential, it provides a consistent benchmark for comparison.

\newpage
\section{Method Hperparameters}
\label{Appendix:method_parameters}
The following is an explanation of the hyperparameters that influenced the running process and effectiveness of our method InfiniRetri when used:
\begin{itemize}
    \item Chunk Size: As shown in Figure 1~\ref{fig:InfiniRetri_Workflow}, in Step 1 (Chunk) of our method, the long document exceeding the context window length of the LLM is first split into several shorter documents of equal length at the boundaries of each sentence. This parameter controls the length of the resulting shorter documents.
    \item Phrase Token Num: As described in Section~\ref{sec:4.2}, our method calculates the importance of tokens retained in the cache based on a phrase-level granularity. This parameter determines the specific number of tokens in each phrase. According to our experimental experience, this parameter is directly related to the length of the correct answer in terms of token count. The retrieval performance is optimized when the parameter setting closely matches the token length of the correct answer.
    \item TopK: As described in Section~\ref{sec:4.2}, our method selects the Top-K sentences containing the highest-scoring tokens based on their importance and stores them in the cache after calculating the importance of each token. This parameter, top-k, directly controls the capacity of the cache, which dynamically changes throughout the execution process of our method.
\end{itemize}

\newpage
\newpage
\section{Example of Visual All Layers Attention Allocation}
\label{appendix:Example of visual all layers}

As described in Section~\ref{sec:observation}, we extracted a Question-Answer pair sample from HotpotQA and fed it into the Qwen2-7B-Instruct to observe and analyze the distribution of attention scores across all layers. 

The Question of QA pair used in the experiment is: \textit{"The FIBT World Championships 1960 took place in a town located in which part of Italy?"}.

The context passages we selected are as follows:\textit{"
Passage 1:
FIBT World Championships 1960
The FIBT World Championships 1960 took place in Cortina d'Ampezzo, Italy for the fifth time. This was an extraordinary event because bobsleigh was not included in the program at the 1960 Winter Olympics in Squaw Valley, California.
Passage 9:
Cortina d'Ampezzo
Cortina d'Ampezzo (......)is a town and comune in the heart of the southern (Dolomitic) Alps in the Province of Belluno, in the Veneto region of Northern Italy. Situated on the Boite river, in an alpine valley, it is a summer and winter sport resort known for its skiing trails, scenery, accommodation, shops and après-ski scene, and for its jet set and Italian aristocratic crowd."} .

Within the context, two sentences are relevant to the Answer. The first is in Passage 1: \textit{"The FIBT World Championships 1960 took place in Cortina d'Ampezzo."}. The second is in Passage 9: \textit{"Cortina d'Ampezzo... is a town and comune in the heart of the southern (Dolomitic) Alps in the Province of Belluno, in the Veneto region of Northern Italy."}. Specifically, \textit{"in the Veneto region of Northern Italy”} is the correct Answer to the Question.

\begin{figure*}[!htbp]
  \centering
  \subfloat[Layer 0\label{fig:all_layer0}]{\includegraphics[width=2.1\columnwidth]{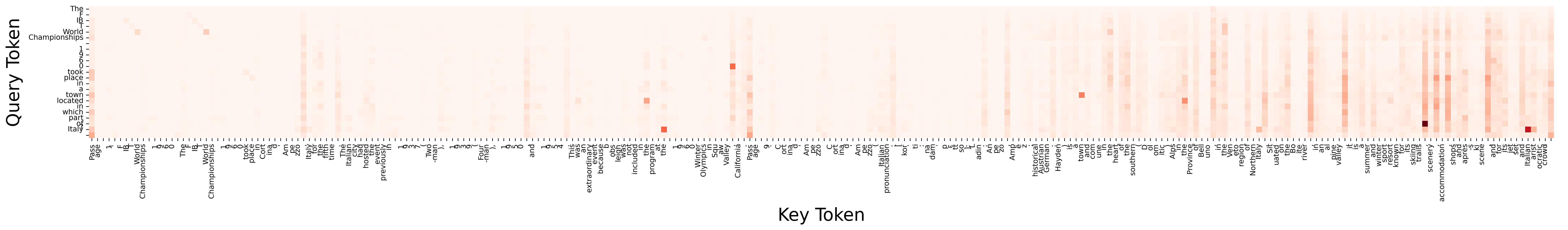}}
  
  \subfloat[Layer 1\label{fig:all_layer1} ]{\includegraphics[width=2.1\columnwidth]{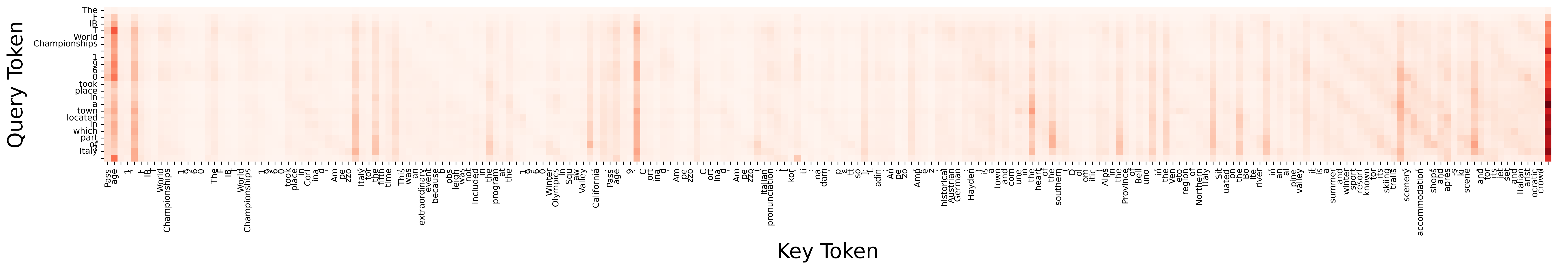}}

  \subfloat[Layer 2\label{fig:all_layer2}]{\includegraphics[width=2.1\columnwidth]{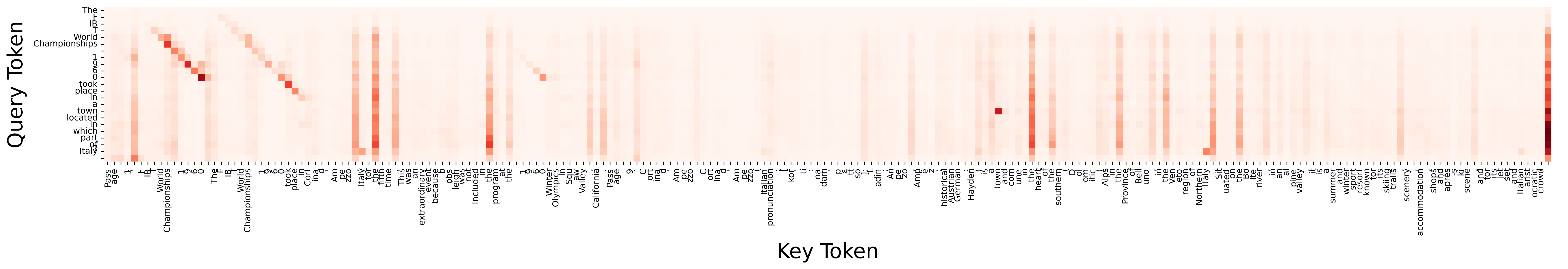}}
  
  \subfloat[Layer 3\label{fig:all_layer3}]{\includegraphics[width=2.1\columnwidth]{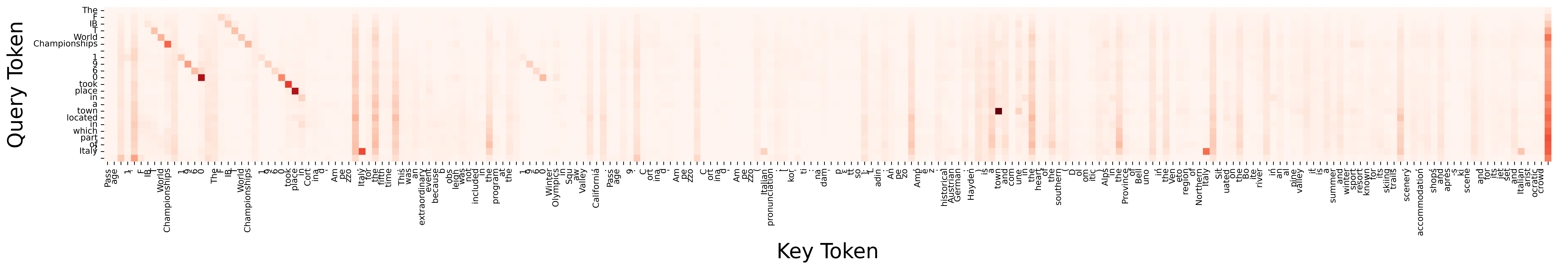}}

  \subfloat[Layer 4\label{fig:all_layer4}]{\includegraphics[width=2.1\columnwidth]{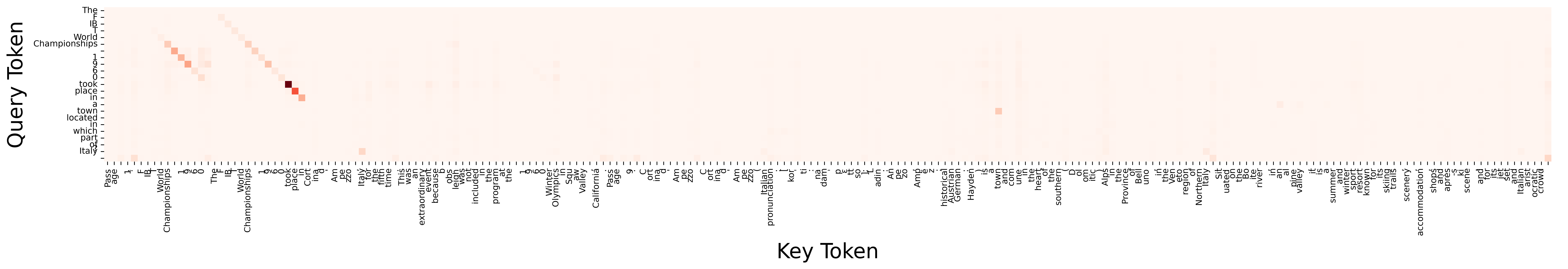}}
  
  \subfloat[Layer 5\label{fig:all_layer5}]{\includegraphics[width=2.1\columnwidth]{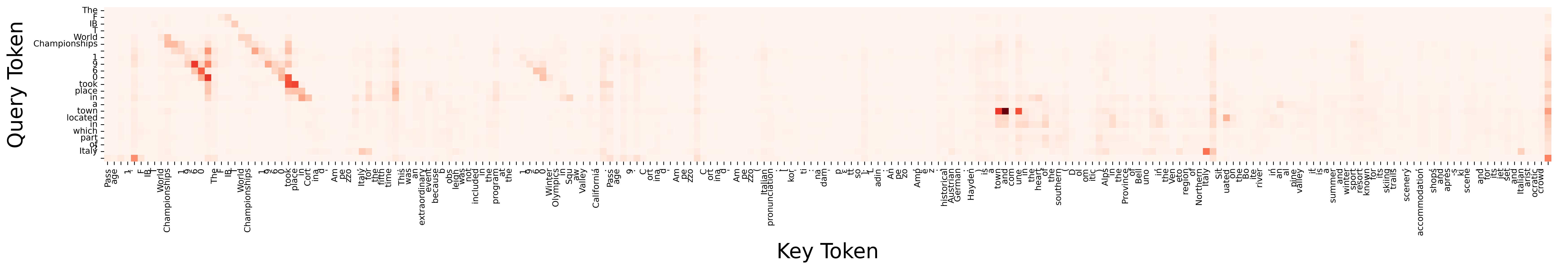}}
  
  \subfloat[Layer 6\label{fig:all_layer6}]{\includegraphics[width=2.1\columnwidth]{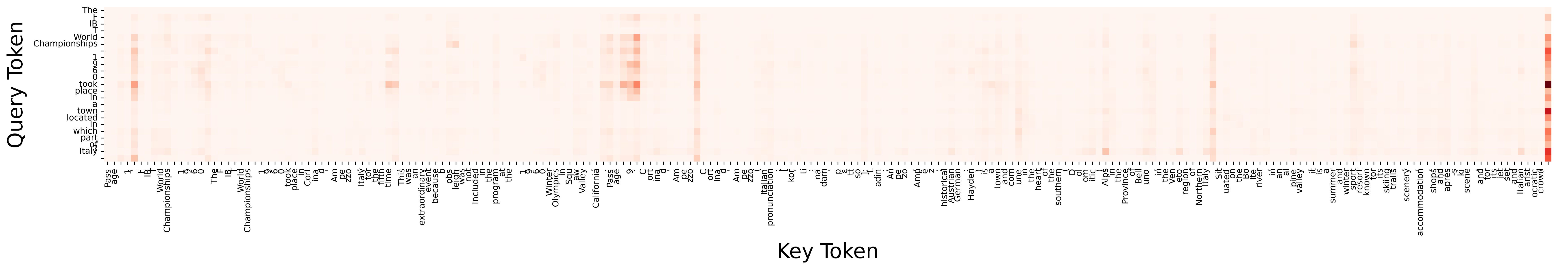}}
  
  \caption{ \label{fig:all_layers_attention_scores0}
    Visual 0-6 layers Attention Scores Heatmap from using Qwen2-7B-Instruct inference in a QA sample segment
  }
\end{figure*}

\begin{figure*}[!htbp]
  \centering
  \subfloat[Layer 7\label{fig:all_layer7}]{\includegraphics[width=2.1\columnwidth]{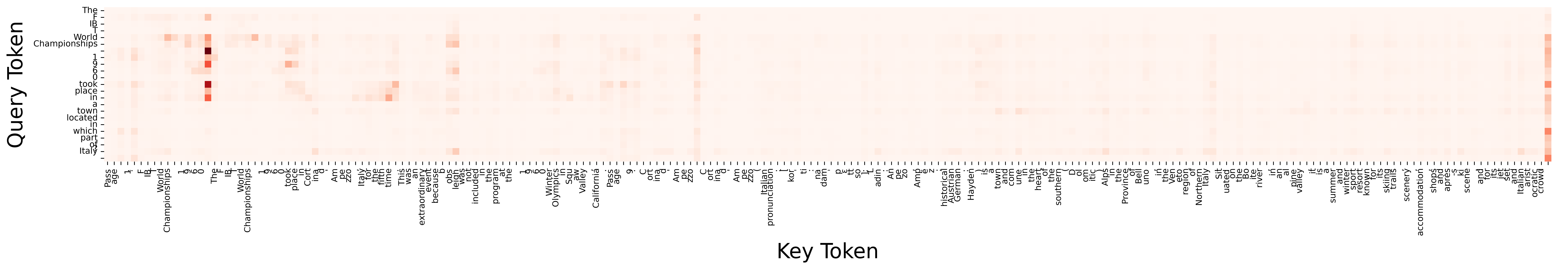}}
  
  \subfloat[Layer 8\label{fig:all_layer8} ]{\includegraphics[width=2.1\columnwidth]{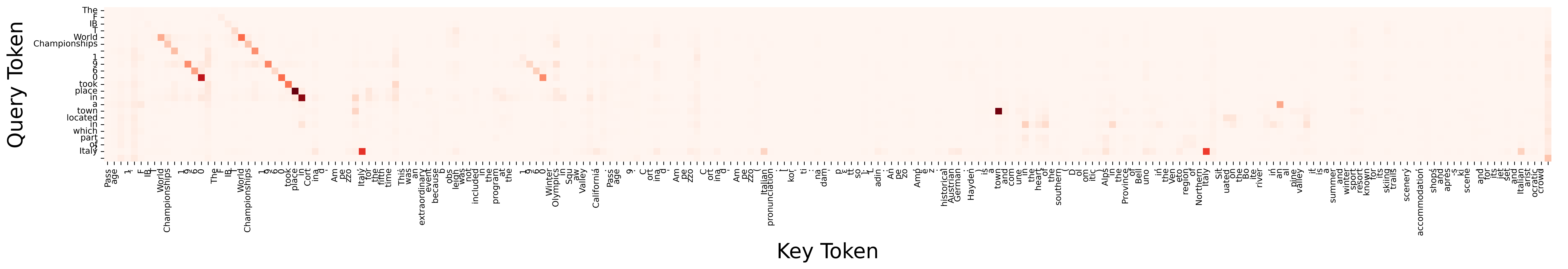}}

  \subfloat[Layer 9\label{fig:all_layer9}]{\includegraphics[width=2.1\columnwidth]{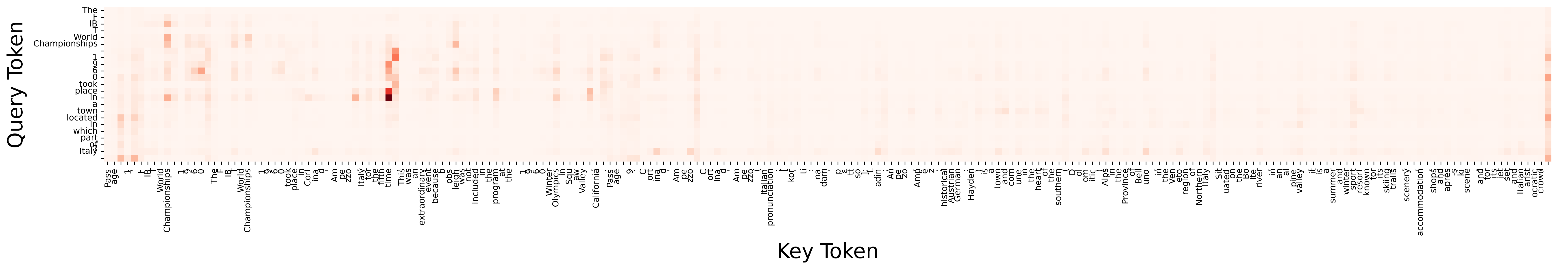}}
  
  \subfloat[Layer 10\label{fig:all_layer10}]{\includegraphics[width=2.1\columnwidth]{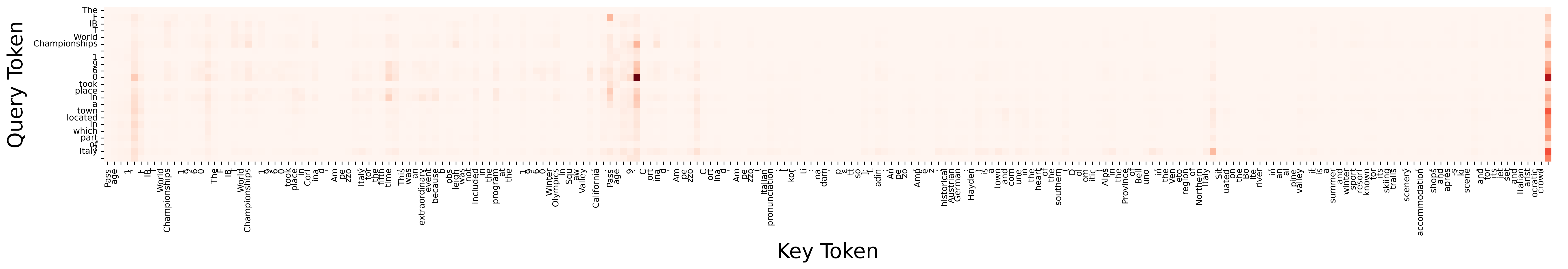}}

  \subfloat[Layer 11\label{fig:all_layer11}]{\includegraphics[width=2.1\columnwidth]{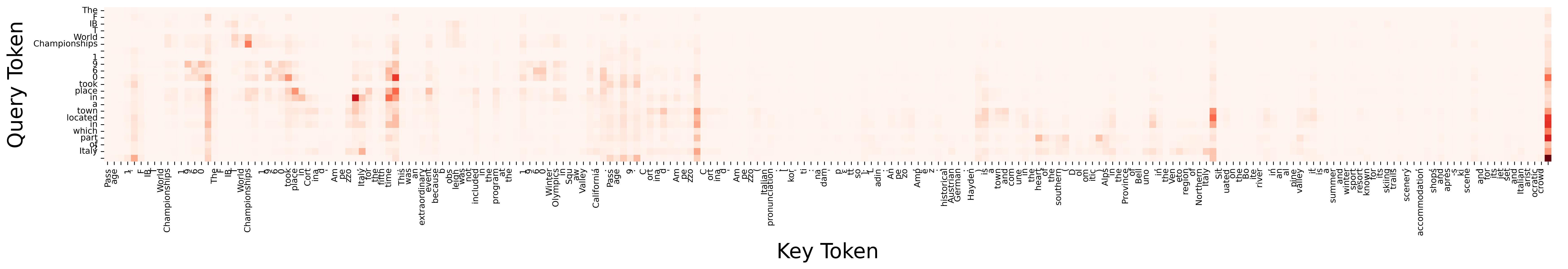}}
  
  \subfloat[Layer 12\label{fig:all_layer12}]{\includegraphics[width=2.1\columnwidth]{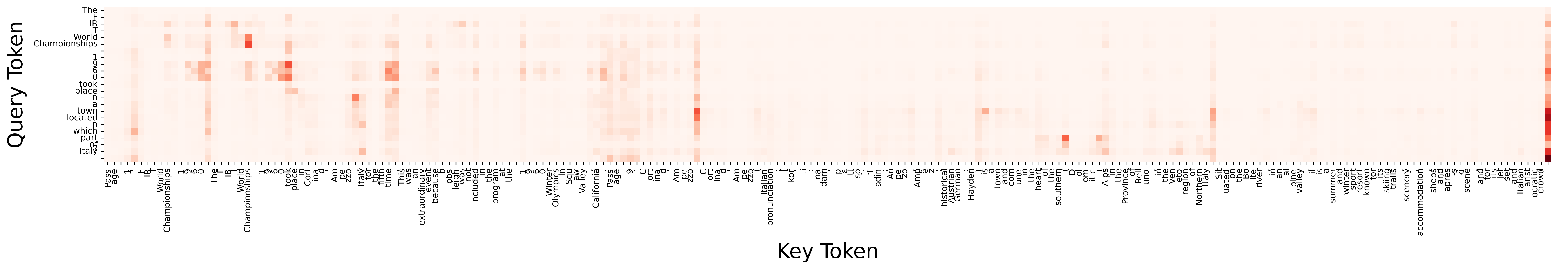}}
  
  \subfloat[Layer 13\label{fig:all_layer13}]{\includegraphics[width=2.1\columnwidth]{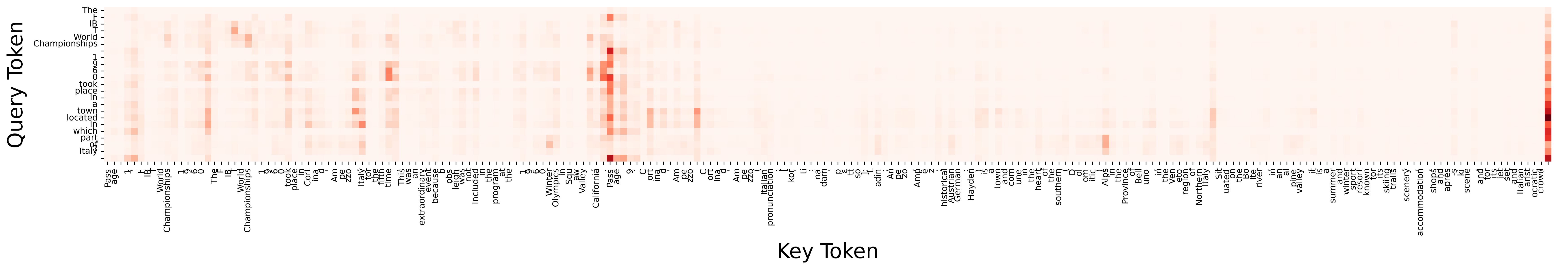}}
  
  \caption{ \label{fig:all_layers_attention_scores1}
    Visual 7-13 layers Attention Scores Heatmap from using Qwen2-7B-Instruct inference in a QA sample segment
  }
\end{figure*}

\begin{figure*}[!htbp]
  \centering
  \subfloat[Layer 14\label{fig:all_layer14}]{\includegraphics[width=2.1\columnwidth]{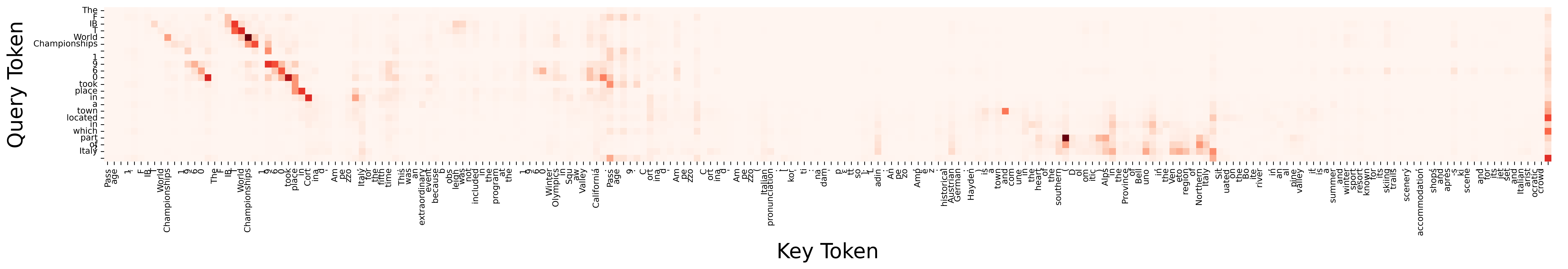}}
  
  \subfloat[Layer 15\label{fig:all_layer15} ]{\includegraphics[width=2.1\columnwidth]{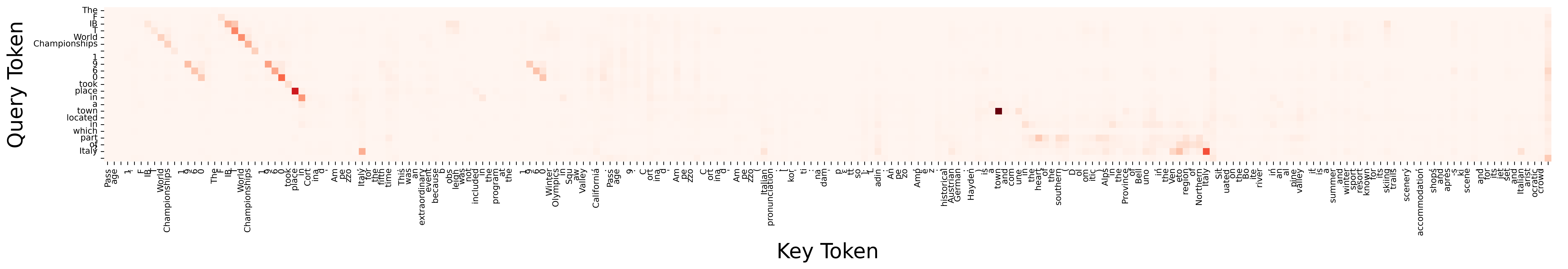}}

  \subfloat[Layer 16\label{fig:all_layer16}]{\includegraphics[width=2.1\columnwidth]{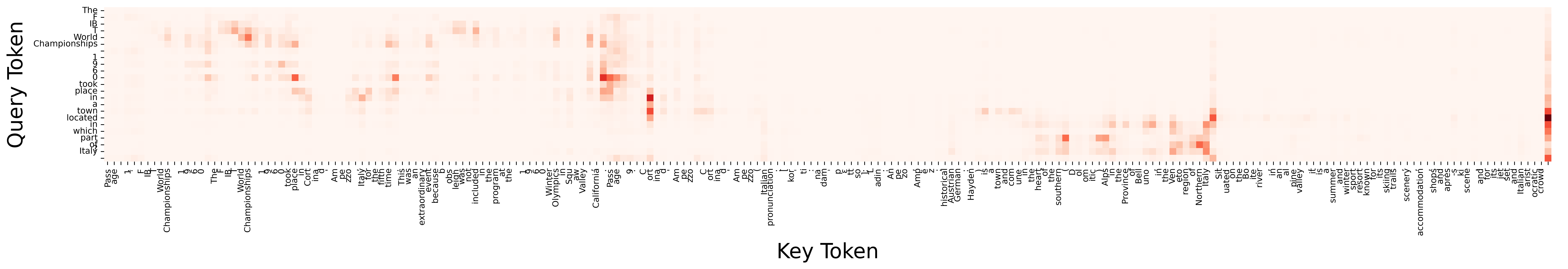}}
  
  \subfloat[Layer 17\label{fig:all_layer17}]{\includegraphics[width=2.1\columnwidth]{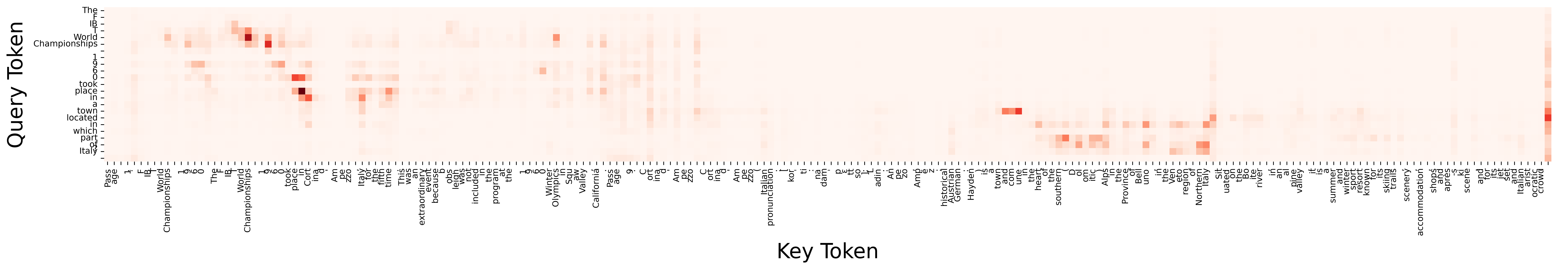}}

  \subfloat[Layer 18\label{fig:all_layer18}]{\includegraphics[width=2.1\columnwidth]{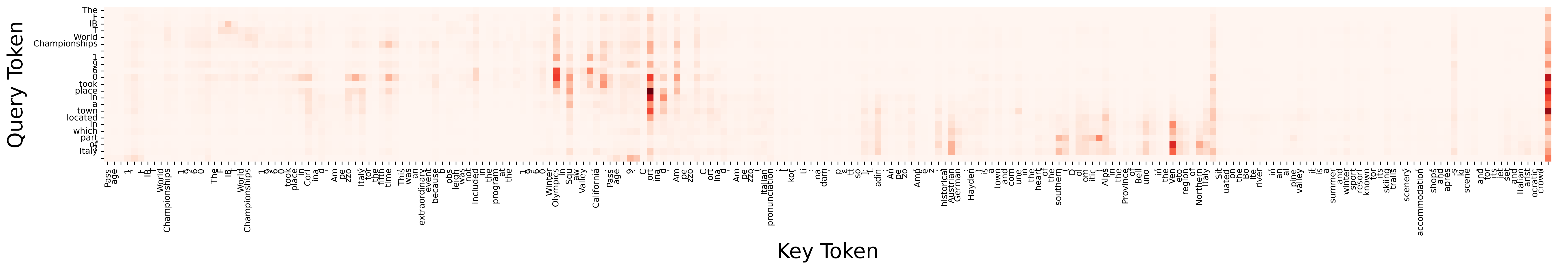}}
  
  \subfloat[Layer 19\label{fig:all_layer19}]{\includegraphics[width=2.1\columnwidth]{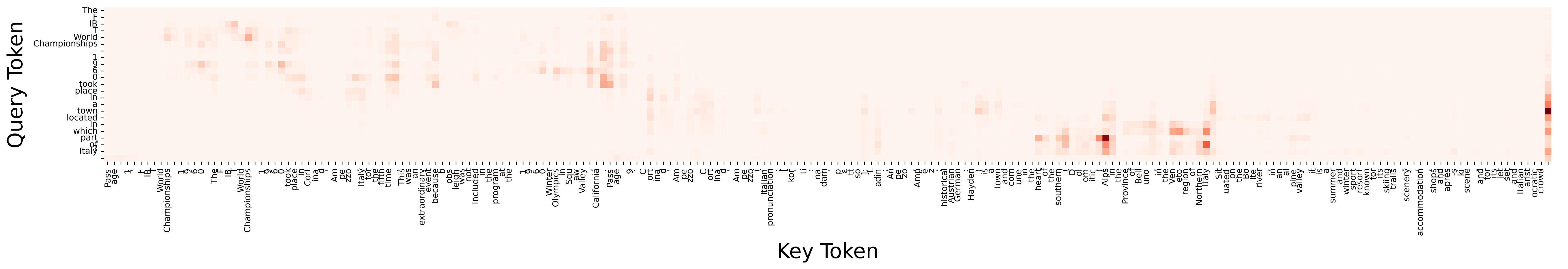}}
  
  \subfloat[Layer 20\label{fig:all_layer20}]{\includegraphics[width=2.1\columnwidth]{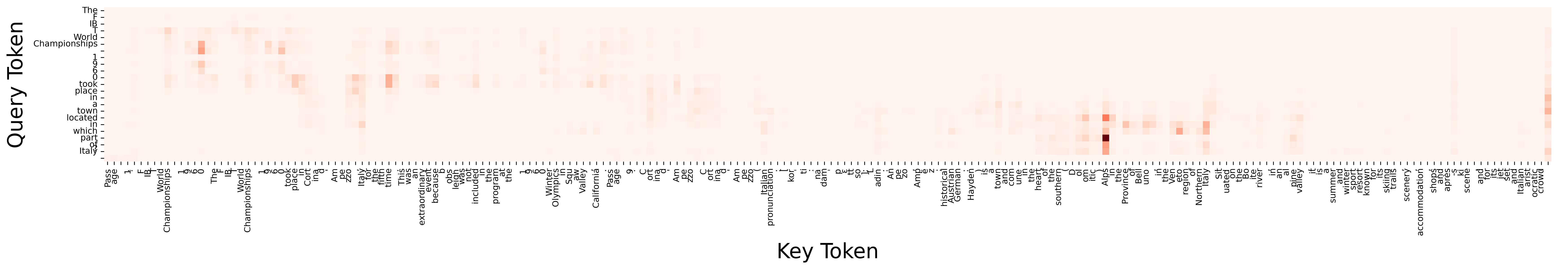}}
  
  \caption{ \label{fig:all_layers_attention_scores2}
    Visual 14-20 layers Attention Scores Heatmap from using Qwen2-7B-Instruct inference in a QA sample segment
  }
\end{figure*}

\begin{figure*}[!htbp]
  \centering
  \subfloat[Layer 21\label{fig:all_layer21}]{\includegraphics[width=2.1\columnwidth]{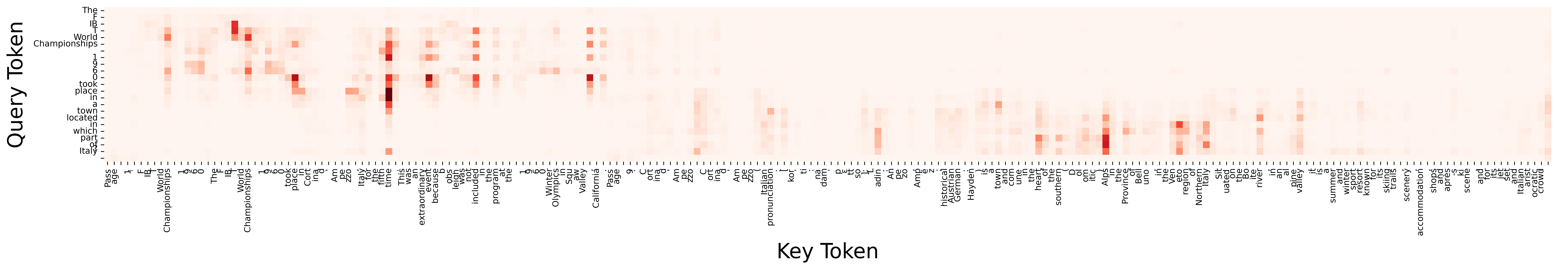}}
  
  \subfloat[Layer 22\label{fig:all_layer22} ]{\includegraphics[width=2.1\columnwidth]{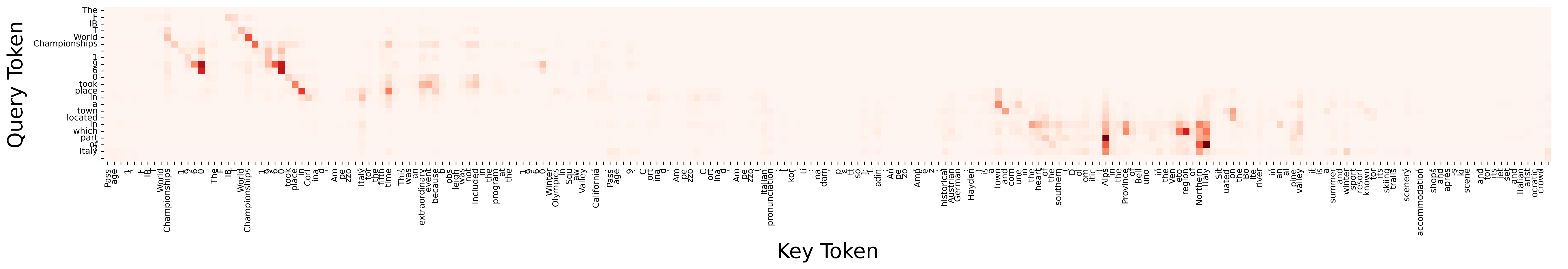}}

  \subfloat[Layer 23\label{fig:all_layer23}]{\includegraphics[width=2.1\columnwidth]{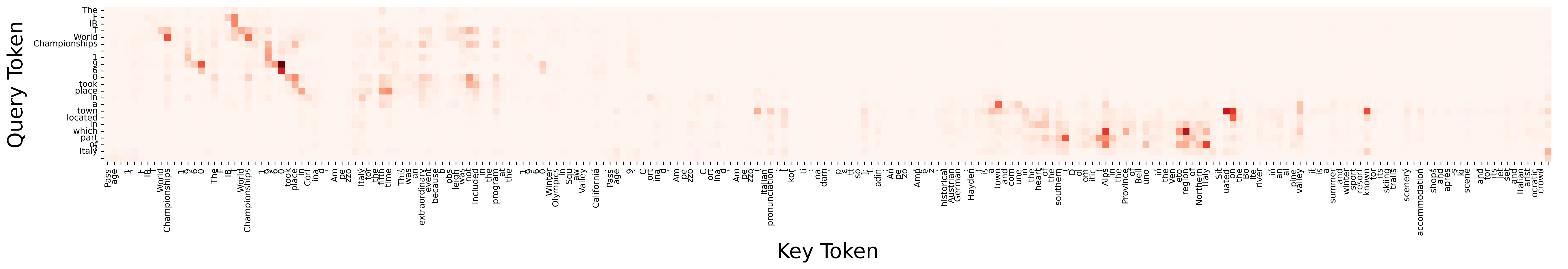}}
  
  \subfloat[Layer 124\label{fig:all_layer24}]{\includegraphics[width=2.1\columnwidth]{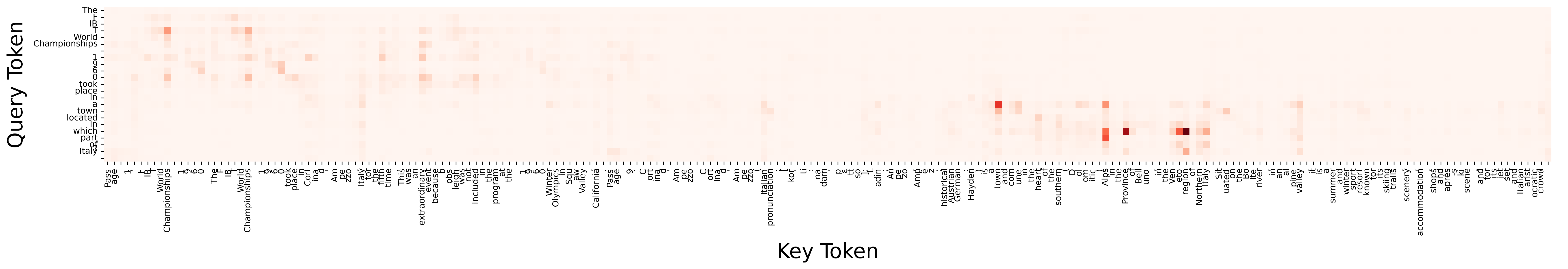}}

  \subfloat[Layer 25\label{fig:all_layer25}]{\includegraphics[width=2.1\columnwidth]{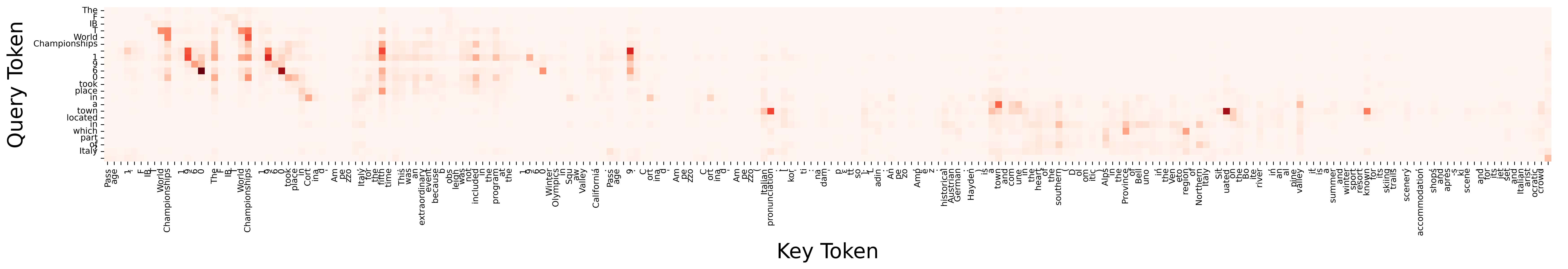}}
  
  \subfloat[Layer 26\label{fig:all_layer26}]{\includegraphics[width=2.1\columnwidth]{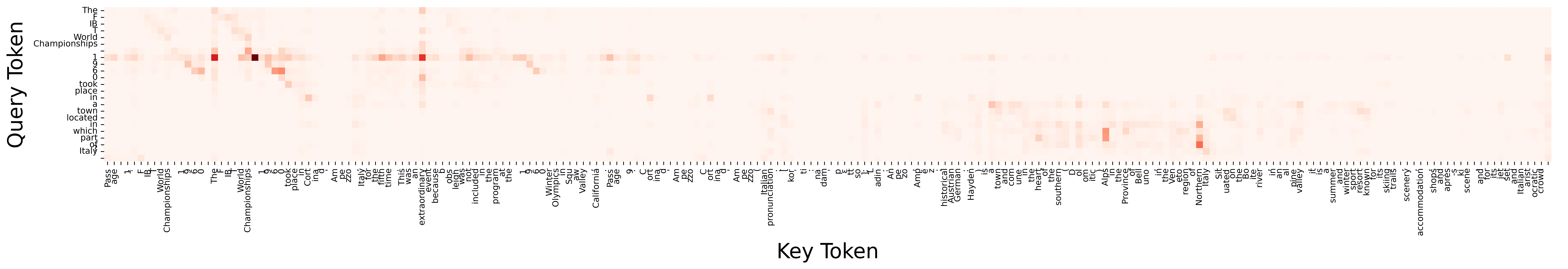}}
  
  \subfloat[Layer 27\label{fig:all_layer27}]{\includegraphics[width=2.1\columnwidth]{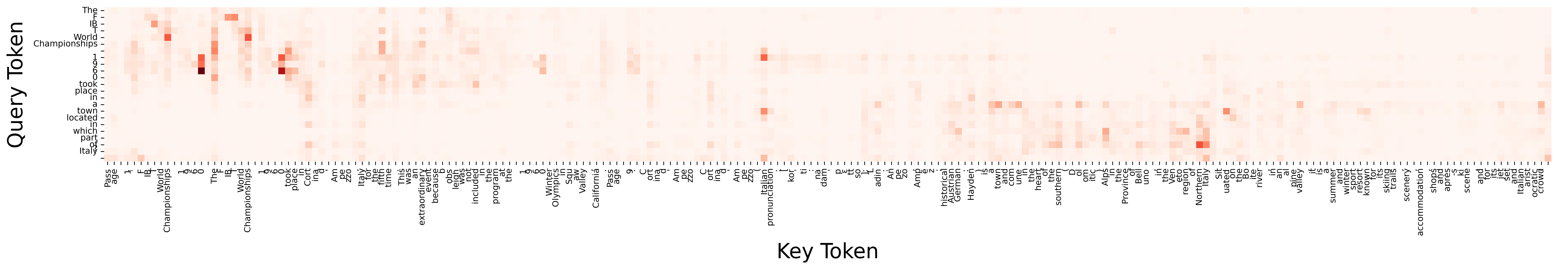}}
  
  \caption{ \label{fig:all_layers_attention_scores3}
    Visual 21-27 layers Attention Scores Heatmap from using Qwen2-7B-Instruct inference in a QA sample segment
  }
\end{figure*}

As shown in Figures~\ref{fig:all_layers_attention_scores0},~\ref{fig:all_layers_attention_scores1},~\ref{fig:all_layers_attention_scores2} and~\ref{fig:all_layers_attention_scores3}, we visualize and present the distribution of attention scores across the 28 layers of this QA sample,  a careful examination of the Query Token to Context Token score distribution reveals that in the attention scores distribution of layers closer to the output in the LLMs, the regions with the highest attention scores correspond precisely to the Answer regions of the QA sample.

\end{document}